\documentclass{bmvc2k}

\usepackage{times}
\usepackage{epsfig}
\usepackage{graphicx}
\usepackage{amsmath}
\usepackage{amssymb}
\usepackage{enumitem}
\usepackage{booktabs}
\usepackage{multirow}

\usepackage{pifont}
\newcommand{\cmark}{\ding{51}}
\newcommand{\xmark}{\ding{55}}
\usepackage{wrapfig}

\usepackage{multirow}

\usepackage{tikz}
\usetikzlibrary{positioning, fit, shapes.geometric, decorations.pathreplacing, calligraphy}

\usepackage{algorithm}
\usepackage{algpseudocode}
\usepackage{capt-of}
\usepackage[export]{adjustbox}  

\usepackage{comment}

\usepackage{colortbl}  %
\usepackage{xcolor}     %

\newcommand{\cbgreen}[1]{%
  \ifmmode %
    \colorbox{green!30!white}{$#1$}%
  \else %
    \colorbox{green!30!white}{#1}
  \fi
}
\newcommand{\cbred}[1]{%
  \ifmmode %
    \colorbox{red!30!white}{$#1$}%
  \else %
    \colorbox{red!30!white}{#1}
  \fi
}
\newcommand{\cbyellow}[1]{%
  \ifmmode %
    \colorbox{yellow!30!white}{$#1$}%
  \else %
    \colorbox{yellow!30!white}{#1}
  \fi
}

\definecolor{lightpink}{RGB}{255, 230, 230}
\definecolor{lightblue}{RGB}{230, 240, 255}
\definecolor{lightyellow}{RGB}{255, 255, 200}

\newcommand{\mname}{DIPSY}

\title{Training-Free Synthetic Data Generation with Dual IP-Adapter Guidance}

\addauthor{Luc Boudier}
{luc.boudier@polytechnique.edu}{$*$ 1}
\addauthor{Loris Manganelli}
{loris.manganelli@polytechnique.edu}{$*$ 1}
\addauthor{Eleftherios Tsonis}
{eleftherios.tsonis@polytechnique.edu}{$*$ 1}
\addauthor{Nicolas Dufour}
{nicolas.dufour@enpc.fr}{1 2}
\addauthor{Vicky Kalogeiton}
{vicky.kalogeiton@polytechnique.edu}{1}

\addinstitution{
 LIX,\\
 École Polytechnique,\\
 IP Paris, CNRS
}
\addinstitution{
 LIGM\\
 École des Ponts,\\
 IP Paris, CNRS, UGE
}

\runninghead{Boudier, Manganelli, Tsonis et al.}{DIPSY: Dual IP-adapter Synthesizer}

\def\eg{\emph{e.g}\bmvaOneDot}

\begin{document}

\maketitle

\begin{figure*}[t]
  \centering
  \includegraphics[width=.79\textwidth]{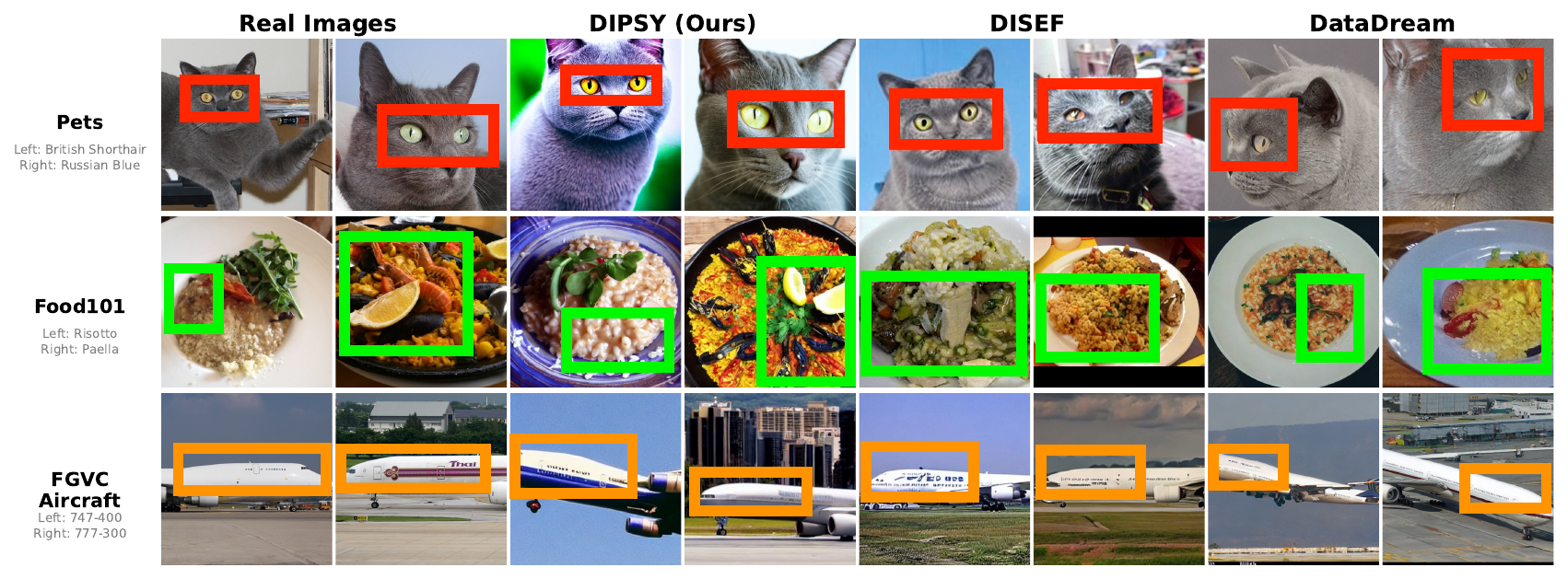}
  \caption{
    Qualitative comparison of synthetic image generation for visually similar class \textbf{pairs} across datasets: British Shorthair vs Russian Blue (Pets), Risotto vs Paella (Food101), and Boeing 747-400 vs 777-300 (FGVC Aircraft). \mname{} generates semantically faithful and visually distinct images, preserving class-specific cues such as \textcolor{red}{eye color} in pets, \textcolor{green}{food-specific textures and toppings}, and \textcolor{orange}{structural aircraft details}. Competing methods (DISEF and DataDream) often produce ambiguous results. Real images included for reference. \textbf{Note:} Colored bounding boxes are manually drawn to emphasize the discriminative features captured by our method.
  }
  \label{fig:qualitative_comparison}
\end{figure*}

\begin{abstract}
Few-shot image classification remains challenging due to the limited availability of labeled examples. Recent approaches have explored generating synthetic training data using text-to-image diffusion models, but often require extensive model fine-tuning or external information sources. We present a novel training-free approach, called \mname, that leverages IP-Adapter for image-to-image translation to generate highly discriminative synthetic images using only the available few-shot examples. \mname{} introduces three key innovations: (1) an extended classifier-free guidance scheme that enables independent control over positive and negative image conditioning; (2) a class similarity-based sampling strategy that identifies effective contrastive examples; and (3) a simple yet effective pipeline that requires no model fine-tuning or external captioning and filtering. Experiments across ten benchmark datasets demonstrate that our approach achieves state-of-the-art or comparable performance, while %
eliminating the need for generative model adaptation or reliance on external tools for caption generation and image filtering.
Our results highlight the effectiveness of leveraging \emph{dual image prompting} with positive-negative guidance for generating class-discriminative features, particularly for fine-grained classification tasks. Project page: \url{https://www.lix.polytechnique.fr/vista/projects/2025_bmvc_dipsy/}.
\end{abstract}

\section{Introduction}
\label{sec:intro}
Few-shot image classification remains challenging when labeled data is scarce or expensive to collect. The ability to accurately categorize images into classes with only a handful of labeled examples has significant implications across domains like medical imaging, industrial inspection, and rare species identification \cite{pachetti2024systematic, liang2023few, lu2024metric}. Traditional approaches to few-shot learning often struggle to extract generalizable representations from limited samples, resulting in models that perform poorly on novel examples \cite{wang2020generalizing}. Generative models \cite{dufour2025dontdropsamplescoherenceaware, podell2023sdxlimprovinglatentdiffusion, rombach2022highresolutionimagesynthesislatent, degeorge2025far} have opened promising avenues to address this limitation by creating synthetic training data to augment the small number of real examples available \cite{azizi2023syntheticdatadiffusionmodels, benigmim2023oneshotunsuperviseddomainadaptation, benigmim2024collaboratingfoundationmodelsdomain, datadream, dacosta2023diversifiedindomainsynthesisefficient}.

Recent work has explored leveraging text-to-image diffusion models to generate synthetic data for training image classifiers. Methods such as DataDream \cite{datadream} adapt generative models through parameter-efficient fine-tuning (e.g. LoRA \cite{hu2021loralowrankadaptationlarge, ruiz2023dreamboothfinetuningtexttoimage}) to align with the target data distribution, achieving impressive results across multiple datasets. Similarly, DISEF \cite{dacosta2023diversifiedindomainsynthesisefficient} employs a strategy that combines in-domain synthesis with efficient fine-tuning of vision-language models. Despite their promising results, these approaches face important limitations: DataDream requires complex model fine-tuning procedures, while DISEF, although training-free for the generative model, relies on captioning and filtering stages and sometimes fails to capture the fine-grained discriminative features necessary for challenging classification tasks. Moreover, on complex datasets with high inter-class similarity, such as FGVC Aircraft, Food, Flowers or Pets, it necessitates the use of very low denoising strengths, which produces images overly close to the originals.

In this paper, we propose DIPSY (\textbf{D}ual \textbf{IP}-Adapter \textbf{Sy}nthesizer), a novel approach that leverages image-to-image translation via IP-Adapter to generate highly discriminative synthetic images for few-shot classification. Unlike DataDream \cite{datadream}, DIPSY is training-free, eliminating computationally expensive generative model adaptation; unlike DISEF \cite{dacosta2023diversifiedindomainsynthesisefficient}, it requires no external captioning or filtering, relying solely on the few-shot examples themselves.
Our framework introduces key innovations including a class similarity-based sampling strategy for selecting effective image prompts and an extended formulation of classifier-free guidance (CFG) that independently modulates positive and negative IP-Adapter guidances. This dual guidance system provides fine-grained control over the generation process, where stronger positive guidance enhances class-specific feature preservation while stronger negative guidance increases inter-class feature discrimination. This allows us to optimize the synthetic data distribution for superior classifier performance.

\mname{} achieves performance comparable to or exceeding state-of-the-art methods on ten few-shot classification benchmarks, particularly excelling in fine-grained classification by accurately capturing discriminative features.
Ablation studies confirm the importance of the negative guidance and class-similarity sampling strategies. By eliminating the need for model fine-tuning, captioning and filtering, \mname{} offers a more accessible solution for real-world applications with limited computational resources or domain-specific knowledge. As illustrated in Fig.~\ref{fig:qualitative_comparison}, our method generates semantically faithful, visually distinct images that preserve class-specific discriminative cues, such as eye color in pets, textures in cuisine, and structural details in aircrafts, while competing methods often produce ambiguous results.

Our contributions are:
(1) A training-free image synthesis framework that avoids external captioning, filtering, and generative model fine-tuning, making it efficient and practical.
(2) A novel extension to classifier-free guidance (CFG) \cite{ho2022classifierfreediffusionguidance} providing independent control over text, positive, and negative image conditioning for precise generation. \mname{} is the first to apply such distinct positive and negative image conditioning within CFG.
(3) A class similarity-based sampling strategy that selects images as negative prompts, enhancing the discriminative power of the generated images.
All these yield a powerful yet simple approach that achieves performance comparable to the state of the art while requiring fewer resources.

\section{Related Work}
\label{sec:related_work}

\noindent \textbf{Generative and Diffusion Models.}
Advances in Variational AutoEncoders (VAEs) \cite{kingma2022autoencodingvariationalbayes} and Generative Adversarial Networks (GANs) \cite{goodfellow2014generativeadversarialnetworks} led to substantial advancements in tasks like high-fidelity image synthesis \cite{brock2019largescalegantraining}, text-to-image synthesis with GANs \cite{kang2023scalingganstexttoimagesynthesis}, image translation \cite{dufour2022scamtransferringhumansimages}, and even audio synthesis \cite{donahue2019adversarialaudiosynthesis}. More recently, diffusion models \cite{song2022denoisingdiffusionimplicitmodels, ho2020denoisingdiffusionprobabilisticmodels, nichol2021improveddenoisingdiffusionprobabilistic} have demonstrated superior performance in generating high-quality and diverse samples, particularly in image synthesis and text-to-image applications \cite{dhariwal2021diffusionmodelsbeatgans, rombach2022highresolutionimagesynthesislatent, podell2023sdxlimprovinglatentdiffusion, pernias2023wuerstchenefficientarchitecturelargescale, saharia2022photorealistictexttoimagediffusionmodels}. Our method leverages IP-Adapter \cite{ipadapter} with diffusion models like Stable Diffusion \cite{rombach2022highresolutionimagesynthesislatent, podell2023sdxlimprovinglatentdiffusion}, integrating image prompts for controllable generation without requiring base model fine-tuning.

\noindent \textbf{Classifier-Free Guidance.}
Controlling generative models often involves conditioning the generation process on input data. Diffusion models can be trained to be unconditional or to incorporate conditioning via architectural mechanisms such as cross-attention \cite{saharia2022photorealistictexttoimagediffusionmodels, rombach2022highresolutionimagesynthesislatent, balaji2023ediffitexttoimagediffusionmodels}, addition to features within network blocks \cite{nichol2021improveddenoisingdiffusionprobabilistic}, or input concatenation. To exert finer control over the generation process or to impose conditions not explicitly learned by the diffusion model during its training, guidance strategies are employed. Two prominent strategies are Classifier Guidance (CG) \cite{dhariwal2021diffusionmodelsbeatgans}, which relies on an external classifier to steer sampling, and Classifier-Free Guidance (CFG) \cite{ho2022classifierfreediffusionguidance}. CFG, which avoids an external classifier by jointly training conditional and unconditional objectives (see Eq.~\eqref{eq:cfg_standard_fancy}) is widely used in text-to-image diffusion models, enabling strong performance in systems like Stable Diffusion \cite{rombach2022highresolutionimagesynthesislatent} (often using CLIP \cite{radford2021learningtransferablevisualmodels}), GLIDE \cite{nichol2022glidephotorealisticimagegeneration}, and Blended Diffusion \cite{Avrahami_2022}.
Notably, InstructPix2Pix \cite{brooks2023instructpix2pixlearningfollowimage} extended CFG for dual conditioning on an image and a text instruction, a principle related to our multi-modal CFG.
Our method takes inspiration from the classical CFG formulation but extends it to handle one textual prompt and two distinct image prompts from IP-Adapter. While recent work has explored CFG refinements such as guidance scheduling \cite{wang2024analysisclassifierfreeguidanceweight}, interval-based guidance \cite{kynkaanniemiapplying}, timestep skipping \cite{dinh2024compress}, or reinterpreting CFG with weaker guiding models \cite{karras2024guiding}, our focus is on adapting the core CFG mechanism for multi-modal image conditioning.

\noindent \textbf{Training with Synthetic Data.}
Recent research has extensively explored training models with synthetic data. This includes augmenting real datasets \cite{bansal2023leavingrealityimaginationrobust, burg2023imageretrievaloutperformsdiffusion, delatolas2025studying, dunlap2023diversifyvisiondatasetsautomatic, zhou2023trainingairimproveimage}, pre-training on synthetic data before fine-tuning on real samples \cite{hammoud2024synthclipreadyfullysynthetic, tian2023stablerepsyntheticimagestexttoimage}, and training entirely on synthetic datasets \cite{hammoud2024synthclipreadyfullysynthetic, sariyildiz2023faketillmakeit, syntheticdatagenerativemodels}. These strategies have been applied to various tasks such as classification \cite{bansal2023leavingrealityimaginationrobust, burg2023imageretrievaloutperformsdiffusion, he2023syntheticdatagenerativemodels, sariyildiz2023faketillmakeit, shipard2023diversitydefinitelyneededimproving}, object detection \cite{lin2023explorepowersyntheticdata}, generative model self-consumption \cite{alemohammad2023selfconsuminggenerativemodelsmad}, and representation learning \cite{tian2023stablerepsyntheticimagestexttoimage}.
To improve synthetic data quality, studies have focused on enhancing faithfulness (e.g., using CLIP filtering \cite{dunlap2023diversifyvisiondatasetsautomatic, he2023syntheticdatagenerativemodels, lin2023explorepowersyntheticdata}, class-based selection \cite{sariyildiz2023faketillmakeit}, or spectral clustering \cite{lin2023explorepowersyntheticdata}) and promoting diversity (e.g., by reducing guidance scales \cite{sariyildiz2023faketillmakeit}, using varied LLM-generated prompts \cite{hammoud2024synthclipreadyfullysynthetic, he2023syntheticdatagenerativemodels}, incorporating diverse domains/backgrounds \cite{dunlap2023diversifyvisiondatasetsautomatic, sariyildiz2023faketillmakeit, shipard2023diversitydefinitelyneededimproving}, or employing prompt templates \cite{burg2023imageretrievaloutperformsdiffusion}). Synthetic data's cost-effectiveness and scalability make it a compelling alternative or supplement to real data  \cite{sariyildiz2023faketillmakeit}.

In the few-shot setting, which is our focus, a key challenge is personalizing large text-to-image models from just a few example images. Methods follow two main paths: prompt-tuning and model fine-tuning. Prompt-tuning methods learn new word embeddings to represent concepts \cite{gal2022imageworthwordpersonalizing, jin2024imageworthmultiplewords}. Fine-tuning methods adjust model weights, either by modifying the entire U-Net with DreamBooth \cite{ruiz2023dreamboothfinetuningtexttoimage} as in DataDream \cite{datadream}, or more efficiently by updating only the cross-attention layers with Custom Diffusion \cite{kumari2023multiconceptcustomizationtexttoimagediffusion}. Our approach aligns with using diffusion models for new training data but distinctively leverages a pre-trained IP-Adapter without fine-tuning the generator, saving considerable training time. Modulating image guidances allows a balance between generation quality and diversity. Furthermore, integrating class similarity to select negative prompts from related classes actively enhances unique class features in generated images, optimizing them for classifier training.

\begin{figure*}[t]  %
    \centering          %
    \vspace{-2mm}
    \begin{tikzpicture}
        \node[draw, dashed, rounded corners, minimum width=3.5cm, minimum height=3.6cm, label=above:DATASET] (dataset) {};
        
        \node at ([xshift=0.55cm,yshift=1.1cm]dataset.west) {CLS i};
        \node[fill=green!50, minimum width=0.5cm, minimum height=0.5cm] at ([xshift=1.4cm,yshift=1.1cm]dataset.west) {};
        \node[fill=green!50, minimum width=0.5cm, minimum height=0.5cm] at ([xshift=2.0cm,yshift=1.1cm]dataset.west) {};
        \node[fill=green!50, minimum width=0.5cm, minimum height=0.5cm] at ([xshift=2.6cm,yshift=1.1cm]dataset.west) {};
        \node[fill=green!50, minimum width=0.5cm, minimum height=0.5cm] at ([xshift=3.2cm,yshift=1.1cm]dataset.west) {};
    
        \node at ([xshift=0.55cm,yshift=-0.3cm]dataset.west) {CLS j};
        \node[fill=red!50, minimum width=0.5cm, minimum height=0.5cm] at ([xshift=1.4cm,yshift=-0.3cm]dataset.west) {};
        \node[fill=red!50, minimum width=0.5cm, minimum height=0.5cm] at ([xshift=2.0cm,yshift=-0.3cm]dataset.west) {};
        \node[fill=red!50, minimum width=0.5cm, minimum height=0.5cm] at ([xshift=2.6cm,yshift=-0.3cm]dataset.west) {};
        \node[fill=red!50, minimum width=0.5cm, minimum height=0.5cm] at ([xshift=3.2cm,yshift=-0.3cm]dataset.west) {};

        \coordinate (arrow_top) at ([xshift=0.5cm, yshift=0.9cm]dataset.west);
        \coordinate (arrow_bottom) at ([xshift=0.5cm, yshift=-0.1cm]dataset.west);

        \draw [<->, gray, thick] (arrow_top) -- node[midway, right, xshift=0.0cm, font=\small] {Similarity} (arrow_bottom);
    
        \node (dots) at ([xshift=-1.4cm,yshift=1.05cm]dataset.south) {$\vdots$};
        \node (dots) at ([xshift=0.5cm,yshift=1.05cm]dataset.south) {$\vdots$};
        
        \node (nshots) at ([xshift=0.5cm,yshift=0.3cm]dataset.south) {n\_shots};
        \draw [
            decorate,
            decoration={
                calligraphic brace,  %
                amplitude=5pt,      %
                raise=-5pt,          %
                mirror              %
            },
            thick                %
        ]
        ([xshift=-0.5cm]nshots.north west) -- ([xshift=0.5cm]nshots.north east); %

        \path[->,line width=1pt, color=green] ([xshift=-0.15cm,yshift=1.0cm]dataset.east)
            edge[bend left] node[above, xshift=0.1cm] {$w_\text{im+}$} +(1,0) node[right] (image1) {};
        \path[->, line width=1pt, color=red] ([xshift=-0.15cm,yshift=-0.3cm]dataset.east)
            edge[bend left] node[below, xshift=0.1cm] {$-w_\text{im-}$} +(1,0) node[right] (image2) {};
        
        \node[above right=-0.6cm and 0.8cm of image1] (img1_node) {\includegraphics[width=1cm]{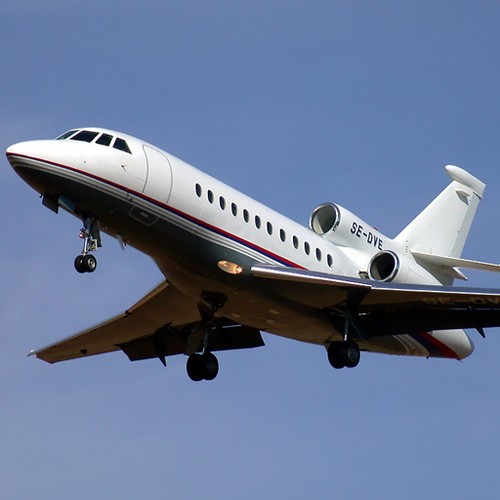}};
        \node[below right=-0.6cm and 0.8cm of image2] (img2_node) {\includegraphics[width=1cm]{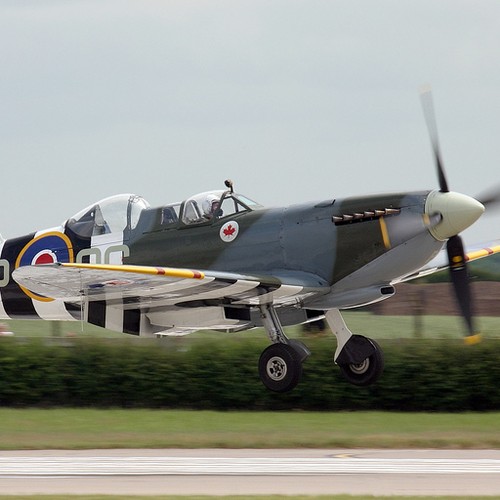}};

        \node[below=0.5] at (img2_node) {Input Images};
        
        \node[fill=green!30!white, minimum width=1.5cm, minimum height=0.7cm, right=0.6cm of img1_node] (ipa1) {IPA+};
        \node[fill=red!30!white, minimum width=1.5cm, minimum height=0.7cm, right=0.6cm of img2_node] (ipa2) {IPA-};
    
        \draw[->, >=stealth, line width=1.0pt, color=green] (img1_node) -- (ipa1);
        \draw[->, >=stealth, line width=1.0pt, color=red] (img2_node) -- (ipa2);

        \begin{scope}[local bounding box=unet_shape]  %
            \coordinate (A) at ([yshift=0.5cm,xshift=0.6cm]ipa1.east);
            \coordinate (B) at ([yshift=-0.5cm,xshift=0.6cm]ipa2.east);
            \coordinate (C) at ([yshift=0.1cm,xshift=2.1cm]ipa2.east);
            \coordinate (F) at ([yshift=-0.1cm,xshift=2.1cm]ipa1.east);
            \coordinate (D) at ([yshift=-0.5cm,xshift=3.6cm]ipa2.east);
            \coordinate (E) at ([yshift=0.5cm,xshift=3.6cm]ipa1.east);
        
            \fill[cyan!30!white] (A) -- (B) -- (C) -- (D) -- (E) -- (F) -- cycle;

        \end{scope}

        \node[below] at (unet_shape.south) {Denoising U-Net};

        \draw[->, >=stealth, line width=1.0pt, color=cyan!30!white] (ipa1.east) -- ([xshift=1.0cm]ipa1.east);
        \draw[->, >=stealth, line width=1.0pt, color=cyan!30!white] (ipa2.east) -- ([xshift=1.0cm]ipa2.east);

        \node[above right=-0.3cm and 0.4cm of unet_shape] (output1) {\includegraphics[width=1cm]{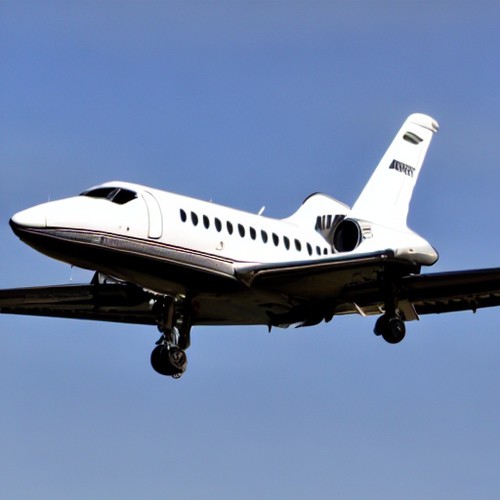}};
        \node[above right=-1.4cm and 0.4cm of unet_shape] (output1) {\includegraphics[width=1cm]{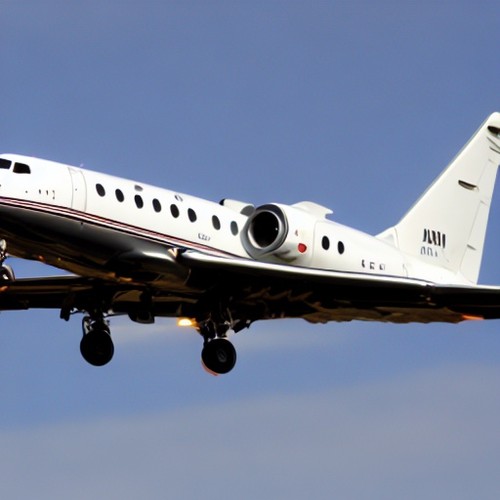}};
        \node[above right=-2.5cm and 0.4cm of unet_shape] (output1) {\includegraphics[width=1cm]{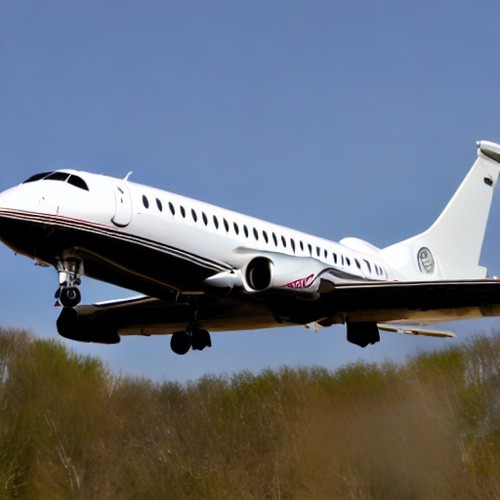}};
        \node[above right=-3.6cm and 0.4cm of unet_shape] (output1) {\includegraphics[width=1cm]{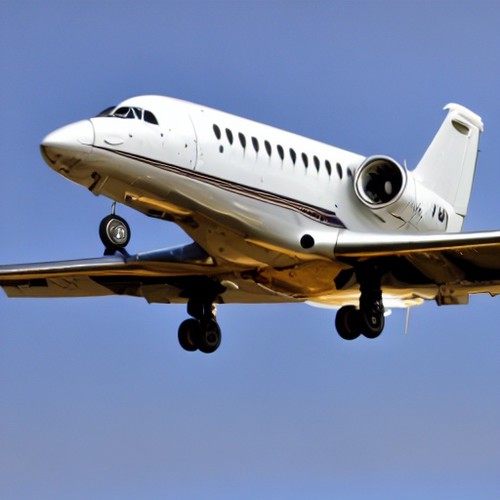}};
        
    \end{tikzpicture}

    \caption{\textbf{Dual IP-Adapter generation pipeline.} An image from the target class (CLS $i$) provides positive conditioning (IPA+, weight $w_\text{im+}$), while an image from a similar class (CLS $j$) provides negative conditioning (IPA-, weight $w_\text{im-}$). These guide the Denoising U-Net to produce images of the target class.} %
    \label{fig:method}      %
\end{figure*}
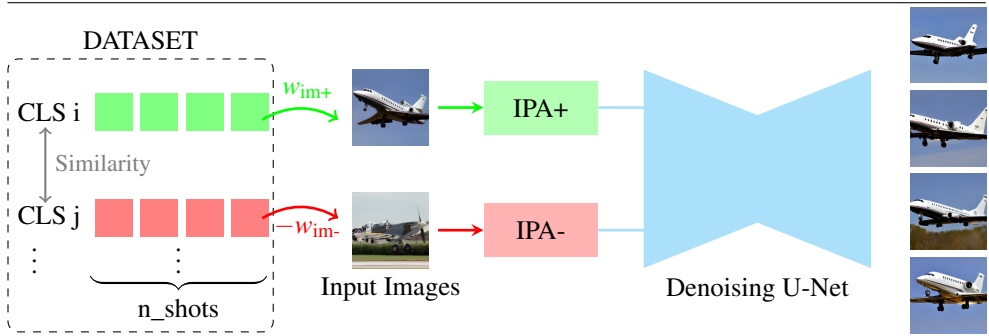

\section{Method}
\label{sec:method}

This section presents our proposed \textbf{D}ual \textbf{IP}-Adapter \textbf{Sy}nthesizer (\mname{}). As Fig.~\ref{fig:method} illustrates, \mname{} generates a synthetic dataset accurately representing target class characteristics from $n_{shots}$ real images by leveraging IP-Adapter \cite{ipadapter} and Stable Diffusion 1.5 \cite{rombach2022highresolutionimagesynthesislatent} (chosen for its compatibility with both DataDream and IP-Adapter). \mname{} introduces a novel extension to classifier-free guidance (CFG) for independent control over textual, positive (green block in Fig.~\ref{fig:method}), and negative (red block in Fig.~\ref{fig:method}) image conditioning (Sec.~\ref{sec:multimodal_cond}). A class similarity-based sampling strategy to select effective negative prompts further enhances this (Sec.~\ref{sec:image_generation}). The resulting synthetic images and original few-shot examples are then used to fine-tune a CLIP model via a weighted loss function (Sec.~\ref{subsec:classification}).

\subsection{Extending Classifier-Free Guidance for Multi-Modal Conditioning}
\label{sec:multimodal_cond}

IP-Adapter with Stable Diffusion 1.5 enables multi-modal conditioning: guidance originates from a textual prompt, a positive image prompt, and a negative image prompt (both using few-shot examples). Standard CFG typically assumes a single conditioning source (text) for scaling the deviation from the unconditional prediction. Our setup requires independent control over text, positive, and negative image conditioning contributions. We achieve this by implementing a modified CFG scheme that permits separate guidance factors for the text prompt, the positive IP-Adapter prompt, and the negative IP-Adapter prompt.

\medskip
To understand our approach, let's first revisit standard classifier-free guidance (CFG). The goal of CFG is to estimate the score function conditioned on a text prompt $c_{\text{text}}$:
\begin{equation}
\epsilon_{\theta}(x_t, c_{\text{text}}) \approx \nabla_{x_t} \log p(x_t | c_{\text{text}}) \quad .
\label{eq:score_text_fancy}
\end{equation}
Here, $\epsilon_{\theta}$ represents the noise prediction model parameterized by $\theta$ at timestep $t$ for noisy input $x_t$. Standard CFG computes the final noise prediction $\hat{\epsilon}_{\theta}$ by extrapolating from the unconditional prediction $\epsilon_{\theta}(x_t)$ towards the text-conditional prediction $\epsilon_{\theta}(x_t, c_{\text{text}})$, scaled by the text guidance factor $w_{\text{text}}$:
\begin{equation}
\hat{\epsilon}_{\theta}(x_t, c_{\text{text}}) = \epsilon_{\theta}(x_t) + w_{\text{text}} (\epsilon_{\theta}(x_t, c_{\text{text}}) - \epsilon_{\theta}(x_t)) \quad .
\label{eq:cfg_standard_fancy}
\end{equation}

Our objective is to extend this framework to incorporate guidance from image prompts in addition to the text prompt, allowing for separate guidance strengths. We first consider the case with one image prompt, $c_{\text{im}}$. We aim to compute the jointly conditioned noise prediction $\hat{\epsilon}_{\theta}(x_t, c_{\text{text}}, c_{\text{im}})$, which corresponds to approximating the score function $\nabla_{x_t} \log p(x_t | c_{\text{text}}, c_{\text{im}})$.
To do so, we first use Bayes' theorem:
\begin{align}
    p(x_t | c_{\text{text}}, c_{im}) &= \frac{p(c_{\text{text}}, c_{\text{im}} | x_t) p(x_t)}{p(c_{\text{text}}, c_{\text{im}})} = \frac{p(c_{\text{text}} | x_t) p(c_{\text{im}} | x_t, c_{\text{text}}) p(x_t)}{p(c_{\text{text}},c_{\text{im}})} \quad .
\end{align}
Following common practice in guided diffusion models, we make an approximation involving guidance parameters $w_{\text{text}}$ and $w_{\text{im}}$:
\begin{align}
    p(x_t | c_{\text{text}}, c_{\text{im}}) &\propto p(c_{\text{text}} | x_t)^{w_{\text{text}}} p(c_{\text{im}} | x_t, c_{\text{text}})^{w_{\text{im}}} p(x_t) \quad ,
\end{align}
Taking the gradient of the log with respect to $x_t$ gives the score function approximation:
\begin{align}
    \nabla_{x_t}\log p(x_t | c_{\text{text}}, c_{\text{im}}) &\approx w_{\text{text}}\nabla_{x_t}\log p(c_{\text{text}} | x_t)
    + w_{\text{im}}\nabla_{x_t}\log p(c_{\text{im}} | x_t, c_{\text{text}}) + \nabla_{x_t}\log p(x_t)\quad . 
    \label{eq:score_approx_fancy}
\end{align}
Now, we apply Bayes' theorem again to express the conditional probabilities in terms of the diffusion model's predictive capabilities:
\begin{equation} \label{eq:bayes_combined_fancy} %
    p(c_{\text{text}} | x_t) = \frac{p(x_t | c_{\text{text}}) p(c_{\text{text}})}{p(x_t)} \quad\text{and}\quad
    p(c_{\text{im}} | x_t, c_{\text{text}}) = \frac{p(x_t | c_{\text{im}}, c_{\text{text}}) p(c_{\text{im}} | c_{\text{text}})}{p(x_t| c_{\text{text}})} \quad .
\end{equation}
By substituting the logarithm gradients derived from \eqref{eq:bayes_combined_fancy} into \eqref{eq:score_approx_fancy}, and assuming the gradients of the log priors $\log p(c_{\text{text}})$ and $\log p(c_{\text{im}} | c_{\text{text}})$ with respect to $x_t$ are negligible, we get:
\begin{align}
\nabla_{x_t}\log p(x_t | c_{\text{text}}, c_{\text{im}}) &\approx \nabla_{x_t}\log p(x_t) +
    w_{\text{text}} \Bigl(\nabla_{x_t}\log p(x_t | c_{\text{text}}) - \nabla_{x_t}\log p(x_t)\Bigl) \notag \\
    &\quad + w_{\text{im}} \Bigl(\nabla_{x_t}\log p(x_t | c_{\text{text}}, c_{\text{im}}) - \nabla_{x_t}\log p(x_t | c_{\text{text}})\Bigl) \quad .
\end{align}
Replacing the score functions $\nabla_{x_t} \log p(x_t | \dots)$ with their corresponding noise predictions $\epsilon_{\theta}(x_t | \dots)$, we arrive at the CFG formula for one text and one image prompt:
{\small
\begin{equation}
\hat{\epsilon}_{\theta}(x_t, c_{\text{text}}, c_{\text{im}}) = \epsilon_{\theta}(x_t) + w_{\text{text}} \Bigl(\epsilon_{\theta}(x_t | c_{\text{text}}) - \epsilon_{\theta}(x_t)\Bigl) 
    + w_{\text{im}} \Bigl(\epsilon_{\theta}(x_t | c_{\text{text}}, c_{\text{im}}) - \epsilon_{\theta}(x_t | c_{\text{text}})\Bigl) \quad . \label{eq:cfg_one_image_fancy}
\end{equation}
}

Finally, we extend this reasoning iteratively to handle two image prompts, $c_{\text{im+}}$ (positive) and $c_{\text{im-}}$ (negative), with distinct guidance scale magnitudes $w_{\text{im+}}$ and $-w_{\text{im-}}$. We apply the logic sequentially: first add the conditioning $c_{\text{im+}}$ with weight $w_{\text{im+}}$ based on $c_{\text{text}}$ (as in Eq.~\eqref{eq:cfg_one_image_fancy}), and then add the conditioning $c_{\text{im-}}$ with weight $-w_{\text{im-}}$ based on the joint conditioning $(c_{\text{text}}, c_{\text{im+}})$. This yields our proposed formula for \mname{}'s dual IP-Adapter guidance.

\medskip %
\noindent\textit{\textbf{\mname{}'s Guidance Scheme}: Given one text prompt $c_{\text{text}}$, one positive image prompt $c_{\text{im+}}$, and one negative image prompt $c_{\text{im-}}$, with corresponding guidance scales $w_{\text{text}}$, $w_{\text{im+}}$, and $-w_{\text{im-}}$, our proposed extended CFG scheme yields the following noise prediction $\hat{\epsilon}_{\theta}$:}

{\footnotesize
\begin{equation}
\begin{split}
    \hat{\epsilon}_{\theta}(&x_t, c_{\text{text}}, c_{\text{im+}}, c_{\text{im-}}) = \epsilon_{\theta}(x_t)
     + w_{\text{text}} \Bigl( \epsilon_{\theta}(x_t | c_{\text{text}}) - \epsilon_{\theta}(x_t) \Bigl)  \\
     &+ \cbgreen{w_{\text{im+}} \Bigl(\epsilon_{\theta}(x_t | c_{\text{text}}, c_{\text{im+}}) - \epsilon_{\theta}(x_t | c_{\text{text}})\Bigl)}
     - \cbred{w_{\text{im-}} \Bigl(\epsilon_{\theta}(x_t | c_{\text{text}}, c_{\text{im+}}, c_{\text{im-}}) - \epsilon_{\theta}(x_t | c_{\text{text}}, c_{\text{im+}})\Bigl)} \quad . \label{eq:cfg_multi_final}
\end{split}
\end{equation}
}

\medskip
\noindent Compared to standard text-only CFG, this extended guidance scheme requires one additional forward pass through the diffusion model's U-Net for the single IP-Adapter case (Eq.~\eqref{eq:cfg_one_image_fancy}), evaluating $\epsilon_{\theta}(x_t | c_{\text{text}}, c_{\text{im+}})$, and two additional forward passes to compute $\epsilon_{\theta}(x_t | c_{\text{text}}, c_{\text{im+}})$ and $\epsilon_{\theta}(x_t | c_{\text{text}}, c_{\text{im+}}, c_{\text{im-}})$, for the dual (positive/negative) IP-Adapter case (Eq.~\eqref{eq:cfg_multi_final}).

\subsection{Image generation}
\label{sec:image_generation}
Given a dataset with only a few images per class (referred to as the number of shots $n_{\text{shots}}$), the number of real images is $n_{\text{real}} = n_{\text{classes}} \cdot n_{\text{shots}}$.
This data is used to generate new synthetic images, with a diffusion model and two IP-Adapters. This number is referred to as $n_{\text{synth}}$.

\vspace{1mm}
\noindent \textbf{Class Similarity.} We model inter-class relationships by computing pairwise CLIP ViT-B/16 cosine similarities from few-shot images, averaging them for class-level scores. A softmax yields a probability distribution for sampling negative classes.
To generate synthetic images, we repeat the process of Fig.~\ref{fig:method} $n_{\text{synth}}$ times: A random real image from class $i$ acts as the positive IP-Adapter prompt.
A negative prompt image is chosen from a different class $j$ based on the similarity distribution.
Both augmented (random crop/rotation) images guide diffusion with weights $w_{\text{im+}}$ (positive) and $-w_{\text{im-}}$ (negative).
This encourages target class features and discourages contrastive ones, aiming for sharper class separation.

\subsection{Classification}
\label{subsec:classification}
We follow the classification approach of DataDream.
A CLIP ViT-B/16 model is fine-tuned on $n_{real}$ real and $n_{synth}$ synthetic images.
For training, each batch $B_i$ contains as many real as synthetic images, replicating real images if necessary. The loss for a given batch is a weighted sum of two Cross Entropy (CE) components as:

{\footnotesize
\begin{equation}
\text{Loss}_i = \lambda \sum_{\substack{j \in B_i \\ \text{ j  real}}} \text{CE} \left( \text{CLIP}(\text{Img}_j), T_j^{\text{real}} \right) + (1 - \lambda) \sum_{\substack{k \in B_i, \\ \text{ k  synthetic}}} \text{CE} \left( \text{CLIP}(\text{Img}_k), T_k^{\text{synth}} \right) \quad .
\label{eq:loss}
\end{equation}
}

\noindent The first component measures how well CLIP classifies real images ($\text{Img}_j$) against their true labels ($T_j^{\text{real}}$). The second one measures how well the model classifies synthetic images ($\text{Img}_k$) against their intended target labels ($T_k^{\text{synth}}$); this target label is set to be the class of the real image used as the positive prompt during the generation of that synthetic image. $\lambda$ (between 0 and 1) controls the relative importance of learning from real vs synthetic images.

\section{Experiments}
\label{sec:experiments}

\subsection{Experimental setup}

\noindent \textbf{Datasets.} Following \cite{datadream}, we use ten benchmarks for few-shot classification, covering a wide range of domains, e.g. textures, objects, scenes, satellite, and fine-grained: Describable Textures Dataset (DTD) \cite{cimpoi14describing}, Food-101 (Food) \cite{bossard14}, Oxford-IIIT Pet Dataset (Pets) \cite{parkhi2012cats}, SUN397 (SUN) \cite{xiao2010sun}, EuroSAT (EuSAT) \cite{helber2019eurosat}, FGVC-Aircraft (AirC) \cite{maji2013fine}, ImageNet (IN) \cite{russakovsky2015imagenet}, OxfordFlowers 102 (FLO) \cite{nilsback2008automated}, StanfordCars (Cars) \cite{krause20133d}, and Caltech-101 (CAL) \cite{fei2006one}. 

\noindent \textbf{Implementation details.} We use  Stable Diffusion 1.5 (SD 1.5) as the generative backbone. For each dataset, we randomly sample few-shot real images, ensuring that smaller shot configurations are strict subsets of the larger ones (\eg, the 8-shot set is a subset of the 16-shot set) to enable fair comparisons. For a fair comparison, in all configurations we follow the setup from the state of the art~\cite{datadream}. We generate 200 images per class, and we use 50 denoising steps.  For the classifier, we use CLIP ViT-B/16 as the base vision-language model and applying LoRA to both the image and text encoders. We set the LoRA rank to 16, $\lambda$ = 0.8, and optimize using AdamW optimizer. We tune the guidance scales for both the text and the two image prompts ($w_{\text{text}}$, $w_{\text{im1}}$,  $-w_{\text{im2}}$) and the learning rate on a per-dataset basis.

\begin{table*}[t]
    \centering
    \resizebox{\textwidth}{!}{%
    \begin{tabular}{ll|cc|cc|cccccccccc|c}
        \toprule
          \textbf{Method} & \textbf{Venue} & \textbf{R} & \textbf{S} & \textbf{\shortstack{Train\\Free}} & \textbf{\shortstack{Captioning\\Filtering-free}} & \textbf{DTD} & \textbf{Food} & \textbf{Pets} & \textbf{SUN} & \textbf{EuSAT} & \textbf{AirC} & \textbf{IN} & \textbf{FLO} & \textbf{Cars} & \textbf{CAL} & \textbf{\shortstack{Avg.\\Acc.}} \\
          \midrule
        Visual Prompt Tuning \cite{jia2022visual} & ECCV22 & \cmark & & & & 66.06 & 86.99 & 91.84 & 70.47 & 92.33 & 36.21 & 69.57 & 90.95 & 69.01 & 95.40 & 66.31 \\
        Text Prompt Tuning \cite{zhou2022learning} & IJCV'22 & \cmark & & & & 70.73 & 83.65 & 89.89 & 72.95 & 87.05 & 45.50 & 67.97 & 97.62 & 81.40 & 95.23 & 79.20 \\
        VPT + TPT & arXiv'23 & \cmark & & & & 72.02 & 83.70 & 88.87 & 72.58 & 90.42 & 48.03 & 67.34 & 98.12 & 81.99 & 95.75 & 79.88 \\
        Classifier Tuning \cite{he2023syntheticdatagenerativemodels} & ICLR23 & \cmark & & & & 73.64 & 87.28 & 92.81 & 76.16 & 87.13 & 46.73 & \cbyellow{73.41} & 86.52 & 82.55 & 96.01 & 80.22 \\
        \midrule
        DISEF \cite{dacosta2023diversifiedindomainsynthesisefficient} & arXiv'23 & \cmark &  \cmark & \cmark & & \cbyellow{75.36} & 87.11 & \cbyellow{94.33} & \cbyellow{77.43} & \cbgreen{94.31} & \cbgreen{63.89} & \cbgreen{73.94} & \cbgreen{98.85} & 88.63 & 96.94 & \cbyellow{85.08} \\
        DataDream \cite{datadream} & ECCV24 & \cmark & \cmark & & \cmark & 73.76 & \cbyellow{87.51} & 93.79 & 76.78 & 92.61 & \cbyellow{63.52} & 72.74 & 98.52 & \cbgreen{90.28} & \cbyellow{98.14} & 84.76 \\
        \mname{} (ours) & & \cmark & \cmark & \cmark & \cmark & \cbgreen{75.55} & \cbgreen{87.81} & \cbgreen{94.73} & \cbgreen{77.78} & \cbyellow{94.13} & 62.35 & {73.24} & \cbyellow{98.56} & \cbyellow{88.95} & \cbgreen{99.18} & \cbgreen{85.23} \\        %
        \bottomrule
    \end{tabular}}
    \caption{\textbf{Performance (\% accuracy) on all datasets with 16-shot real data.} 
                Columns "R" and "S" indicate whether the method uses few-shot real data and synthetic data, respectively. 
                Top section shows results from PEFT-based methods using only real data. 
                Bottom section includes methods that generate synthetic data and use it alongside few-shot real data. 
                "Train-free" denotes methods that require no training prior to generation, while "Captioning Filtering-free" refers to those that do not rely on external models for captioning or filtering during the generation pipeline. 
                DataDream results were reproduced using the authors’ released code, adapted to use a Stable Diffusion 1.5 backbone for consistency with other methods. 
                \cbgreen{Best} and \cbyellow{second best} results are highlighted.
                 }
    \label{tab:results}
\end{table*}

\subsection{Comparison to the state of the art}

\paragraph{Quantitative results.}
We compare \mname{} with two groups of methods: (i) few-shot parameter efficient fine-tuning (PEFT) approaches that use no synthetic data for tuning, and (ii) methods that generate synthetic data and use them, along with few-shot real data, to fine-tune. For consistency, we report baseline results (VPT, TPT, Classifier Tuning, and DISEF), taken directly from DISEF~\cite{dacosta2023diversifiedindomainsynthesisefficient}, where all methods are evaluated under the same few-shot setting. We reproduce DataDream using its official implementation, adapted to SD1.5. 

Tab.~\ref{tab:results} reports the top-1 classification accuracy of all methods in all 10 datasets. \mname{} achieves the highest average accuracy overall, despite being entirely training-free and not relying on external captioning and filtering. It establishes new SOTA performance on 5 datasets and ranks within the top-2 on 8, including challenging benchmarks such as the fine-grained DTD and the satellite imagery EuroSAT. In the remaining 2 datasets, \mname{} results in competitive performance. Our method shows relative limitations on datasets like FGVC-Aircraft, where the high inter-class similarity and the limited discriminative capacity of the diffusion backbone drop its performance. In contrast, the training-based DataDream benefits from learning class-specific features, thus better separating visually similar categories. Overall, \mname{} demonstrates strong generalization and competitive accuracy across diverse visual domains, without the need for fine-tuning or extra supervision.

\begin{table*}[t]
\begin{minipage}[t]{\textwidth}
    \centering
    \resizebox{\textwidth}{!}{%
    \begin{tabular}{l|cccccccccc|c}
        \toprule
        \textbf{Method} & \textbf{DTD} & \textbf{Food} & \textbf{Pets} & \textbf{SUN} & \textbf{EuSAT} & \textbf{AirC} & \textbf{IN} & \textbf{FLO} & \textbf{Cars}  & \textbf{CAL} & \textbf{\shortstack{Average\\Accuracy}} \\
        \midrule
        SD1.5 (no adapters) & 74.41 & \cbyellow{87.74} & \cbyellow{94.58} & 77.41 & 92.89 & \cbgreen{63.44} & \cbyellow{73.21} & 98.21 & \cbyellow{88.92}  & 98.98 & \cbyellow{84.98} \\
        \mname{} (empty text prompt) & \cbyellow{74.42} & 87.60 & 94.57 & \cbyellow{77.42} & \cbyellow{93.58} & 61.68 & 72.68 & \cbyellow{98.53} & 87.29  & \cbyellow{99.16} & 84.69 \\
        \mname{} (ours) & \cbgreen{75.55} & \cbgreen{87.81} & \cbgreen{94.73} & \cbgreen{77.78} & \cbgreen{94.13} & \cbyellow{62.35} & \cbgreen{73.24} & \cbgreen{98.56} & \cbgreen{88.95}  & \cbgreen{99.18} & \cbgreen{85.23} \\
        \bottomrule
    \end{tabular}}
    \caption{\textbf{Comparison (\% accuracy) between our method and two baselines}: (1) Stable Diffusion 1.5 (SD1.5) without image prompts, and (2) our method without text prompts.}  
    \label{tab:baselines}
    
\end{minipage}
    \begin{minipage}[t]{0.35\textwidth}
        \centering
        \resizebox{\linewidth}{!}{%
        \begin{tabular}{c|c|c}
            \toprule
            \textbf{\shortstack{Number\\of Classes}} & \textbf{\shortstack{\mname{}\\(ours)}} & \textbf{DataDream} \\
            \midrule
            10   & 1s      & 36m \\
            100  & 57s     & 6h \\
            1000 & 8m      & 61h \\
            \bottomrule
        \end{tabular}}
        \caption{\textbf{Precompute time comparison.} Runtime for precomputing using \mname{} and DataDream on a single H100 GPU with 16 shots.}
        \label{tab:precompute}
    \end{minipage}
    \hfill
    \begin{minipage}[t]{0.63\textwidth}
        \centering
        \resizebox{\linewidth}{!}{%
        \begin{tabular}{ccc|cccc}
            \toprule
            \textbf{\shortstack{Negative\\Image Prompt}} & \textbf{\shortstack{Class\\Similarity}} & \textbf{\shortstack{Data\\Augmentation}} & \textbf{DTD} & \textbf{Pets} & \textbf{EuSAT} & \textbf{AirC} \\
            \midrule
            \xmark & \xmark & \xmark & 75.02 & 94.72 & 93.44 & 59.70 \\
            \xmark & \xmark & \cmark & 74.70 & {94.57} & 92.74 & 60.13 \\
            \xmark & \cmark & \xmark & \multicolumn{4}{c}{\multirow{2}{*}{Class Sim. only defined for negative.}} \\
            \xmark & \cmark & \cmark & \\
            \cmark & \xmark & \xmark & 75.22 & \cbgreen{94.79} & 93.36 & 61.54 \\
            \cmark & \xmark & \cmark & \cbyellow{75.36} & 94.70 & 93.23 & 60.49 \\
            \cmark & \cmark & \xmark & 74.93 & \cbyellow{94.77} & \cbyellow{93.49} & \cbyellow{61.55} \\
            \cmark & \cmark & \cmark & \cbgreen{75.55} & {94.73} & \cbgreen{94.13} & \cbgreen{62.35} \\
            \bottomrule
        \end{tabular}}
        \caption{\textbf{\mname{} component ablation}: \textit{Negative Image Prompt} introduces contrastive guidance; \textit{Class Similarity} selects challenging negatives; and \textit{Data Augmentation} adds variability to prompt inputs. All components contribute to improved performance, while the full setup yields the best results.}
        \label{tab:components}
    \end{minipage}
\end{table*}

\begin{table}[t]
    \centering
    \resizebox{0.8\linewidth}{!}{%
    \begin{tabular}{l|cccccc}
        \toprule
        \textbf{Method} & \textbf{\shortstack{Prep Time\\(1000 classes)}} & \textbf{\shortstack{Gen. Time\\per Image}} & \textbf{\shortstack{Gen. Time\\per Class}} & \textbf{\shortstack{Denoising\\Steps}} \\
        \midrule
        DISEF \cite{dacosta2023diversifiedindomainsynthesisefficient} & 0 & 0.80s & 160s & 15 \\
        DataDream \cite{datadream} & 61h & 1.36s & 271s & 50 \\
        \mname{} (single IP-Adapter) & 0 & 2.59s & 518s & 50 \\
        \mname{} (ours) & 8m & 3.91s & 781s & 50 \\
        \bottomrule
    \end{tabular}}
    \caption{\textbf{Pipeline efficiency comparison for 1000 classes.} While \mname{} requires additional forward passes per diffusion step, it eliminates training overhead and external dependencies. Single IP-Adapter version offers faster generation while maintaining training-free operation. Experiments run on H100 GPU.
    }

    \label{tab:efficiency_comparison}
\end{table}

\paragraph{Qualitative results.}
Comparing \mname{} with DISEF and DataDream on challenging, visually similar class pairs (Fig.~\ref{fig:qualitative_comparison}) demonstrates its superior ability to capture critical discriminative features and generate high-fidelity images. For instance, in the Pets dataset, \mname{} accurately renders distinct eye colors (e.g., orange for British Shorthairs, green-gray for Russian Blues), details DISEF inverts and DataDream entirely misses. In Food101, \mname{} not only successfully differentiates paella from risotto by highlighting typical seafood toppings where other methods produce ambiguous dishes but also, as further illustrated for Food101 classes like Cupcakes and Steak (Fig.~\ref{fig:grouped_by_method_food}), captures defining visual characteristics and maintains aesthetic consistency by generating realistic textures and compositions. Furthermore, for aircraft, \mname{} captures the Boeing 747-400's characteristic upper-deck bump, a key structural cue absent in DISEF’s generation. These examples underline \mname{}'s robustness in producing class-distinctive, visually plausible synthetic samples crucial for fine-grained, few-shot regimes.

\paragraph{Efficiency.}
To provide a comprehensive efficiency analysis, we present two comparisons that reveal the trade-offs between different approaches. Tab.~\ref{tab:precompute} shows the precompute time required by \mname{} and DataDream, demonstrating our method's scalability advantage. Since \mname{} is entirely training-free, the only computation involved is building a similarity matrix between classes, a lightweight operation. In contrast, DataDream requires training a separate LoRA module for each class, resulting in significantly higher computational cost.

A pipeline comparison \ref{tab:efficiency_comparison} shows that while DIPSY's dual IP-Adapter requires more forward passes, it eliminates training overhead and external dependencies. In contrast, DISEF is faster due to fewer noising steps, but this reduces the variability of generated images. The single IP-Adapter version of DIPSY offers a faster, training-free alternative with competitive performance as seen in Tab.~\ref{tab:components} (line 2). This makes DIPSY more practical for immediate deployment, avoiding training failures and external dependencies despite a higher per-image computational cost.

\subsection{Analysis and ablations}

\noindent \textbf{Impact of conditioning}. Tab.~\ref{tab:baselines} presents a comparison between \mname{} and two ablation baselines: SD1.5 without image prompts, and \mname{} with an empty text prompt. The complementary role of multimodal conditioning in \mname{} consistently improves results, enabling better alignment with class semantics and visual characteristics.

\paragraph{Ablation of \mname{} components.}

An ablation study on DIPSY's components, detailed in Tab.~\ref{tab:components}, confirms the value of our positive-negative guidance. This strategy pairs a random positive image prompt from the target class with a negative prompt from a visually similar class, determined by CLIP feature similarities. The model is consequently pushed to learn subtle, discriminative features. While the single IP-Adapter with only positive guidance is effective, adding a negative prompt signficantly enhances inter-class feature discrimination and boosts performance, especially for fine-grained tasks.

\begin{figure*}[t]
\begin{minipage}[t]{\textwidth}
  \centering
  \setlength{\tabcolsep}{2pt}
  \resizebox{\linewidth}{!}{%
  \begin{tabular}{cc}
    \begin{tabular}{c}
      \textbf{Real Images} \\[1mm]
      \textit{Cupcakes} \\
      \includegraphics[valign=m,width=0.22\linewidth]{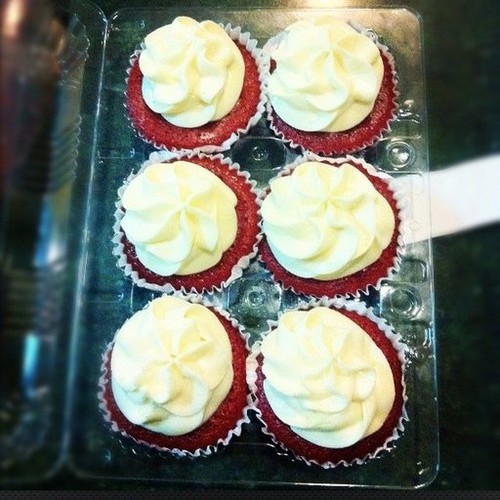} 
      \includegraphics[valign=m,width=0.22\linewidth]{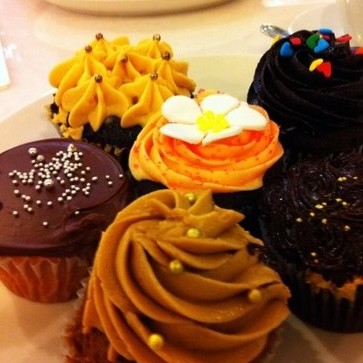} 
      \includegraphics[valign=m,width=0.22\linewidth]{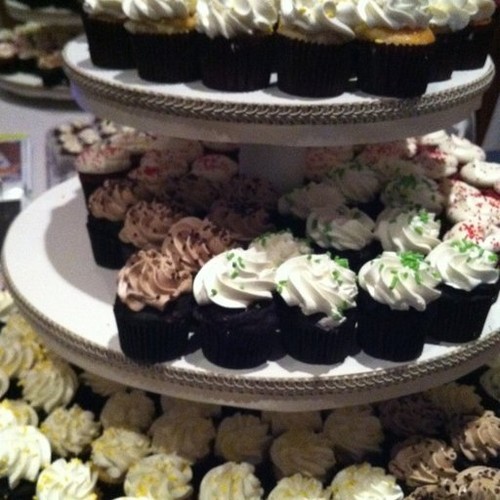} 
      \includegraphics[valign=m,width=0.22\linewidth]{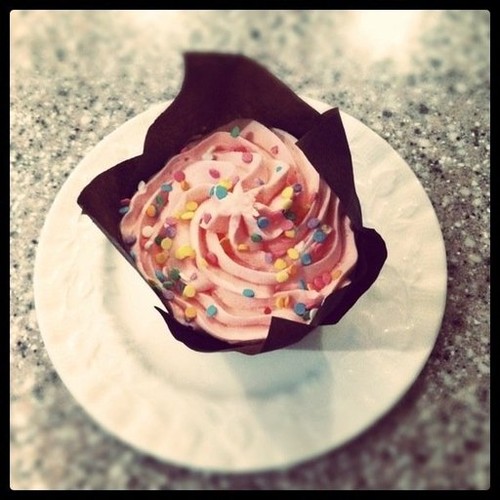} \\
      \textit{Steak} \\
      \includegraphics[valign=m,width=0.22\linewidth]{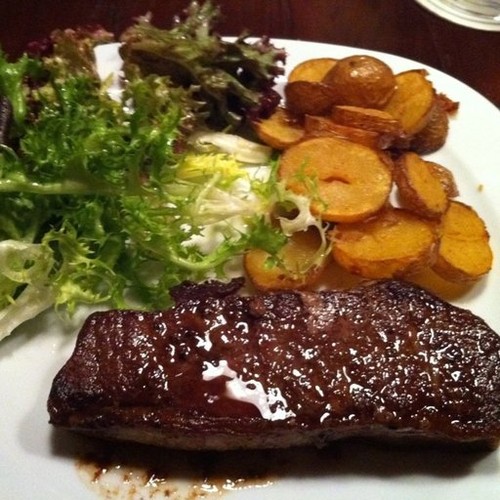} 
      \includegraphics[valign=m,width=0.22\linewidth]{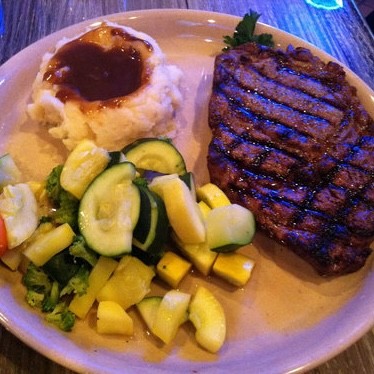} 
      \includegraphics[valign=m,width=0.22\linewidth]{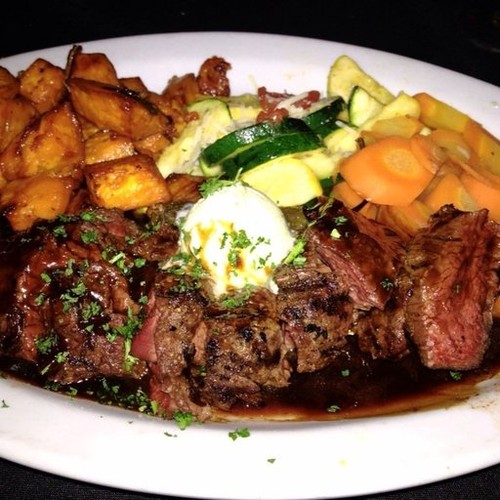} 
      \includegraphics[valign=m,width=0.22\linewidth]{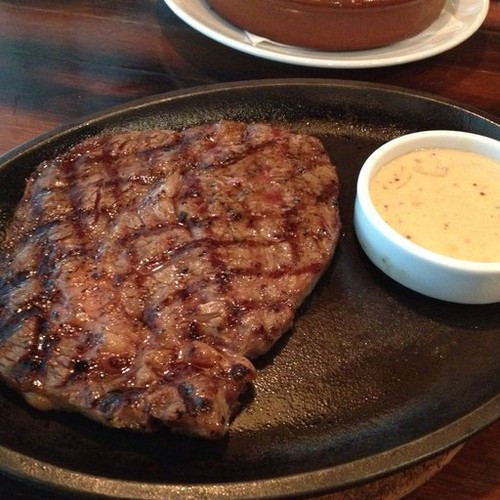} \\
    \end{tabular}
    &
    \begin{tabular}{c}
      \textbf{DIPSY (ours)} \\[1mm]
      \textit{Cupcakes} \\
      \includegraphics[valign=m,width=0.22\linewidth]{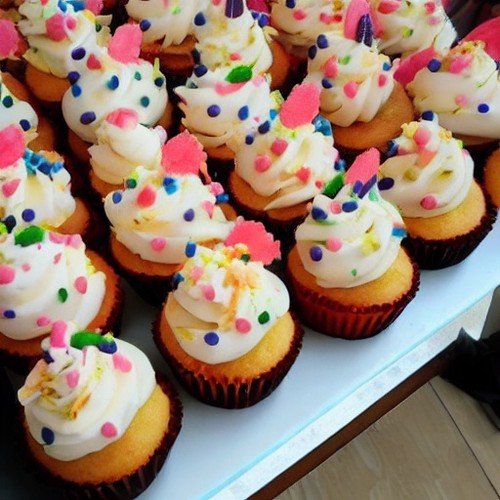} 
      \includegraphics[valign=m,width=0.22\linewidth]{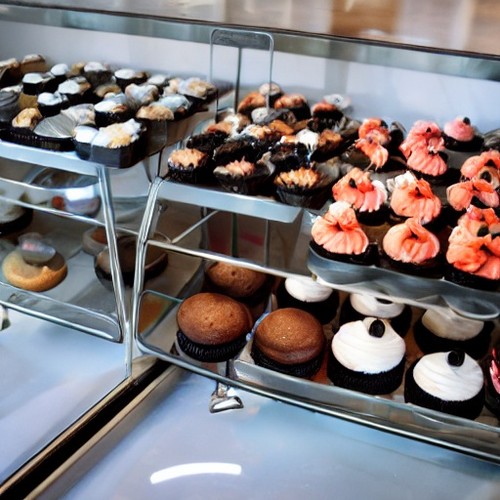} 
      \includegraphics[valign=m,width=0.22\linewidth]{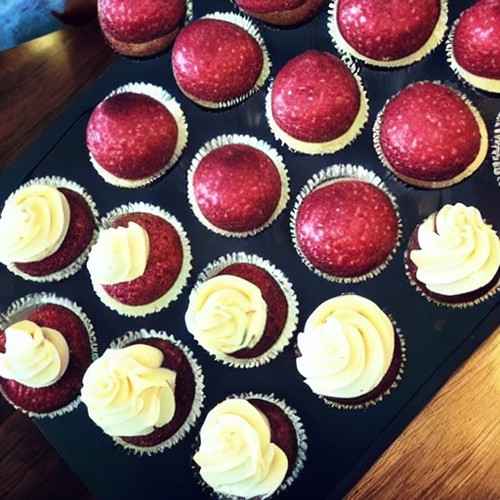} 
      \includegraphics[valign=m,width=0.22\linewidth]{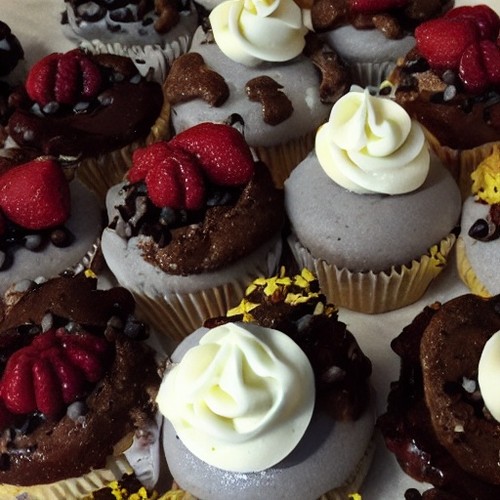} \\
      \textit{Steak} \\
      \includegraphics[valign=m,width=0.22\linewidth]{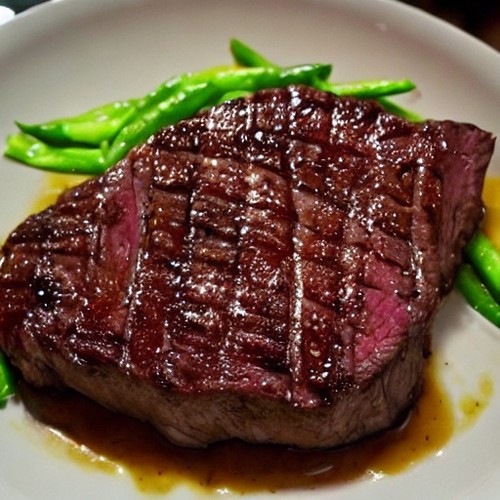} 
      \includegraphics[valign=m,width=0.22\linewidth]{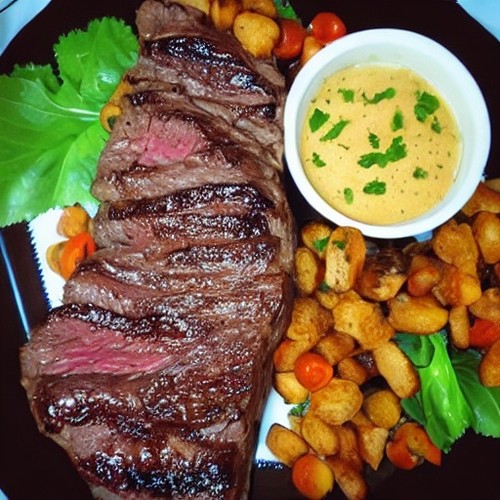} 
      \includegraphics[valign=m,width=0.22\linewidth]{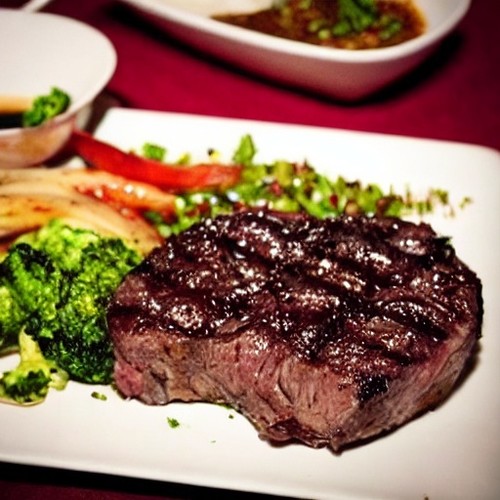} 
      \includegraphics[valign=m,width=0.22\linewidth]{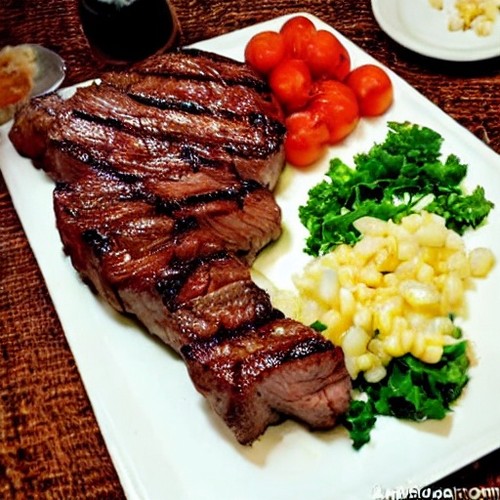} \\
    \end{tabular}
    \\
  \end{tabular}}
  \captionof{figure}{
    Qualitative comparison of real and DIPSY generated synthetic images from the classes Cupcakes and Steak from the Food101 dataset.
  }
  \label{fig:grouped_by_method_food}
  \end{minipage}
  \centering
  \begin{minipage}[t]{0.48\linewidth}
    \centering
    \includegraphics[width=\linewidth]{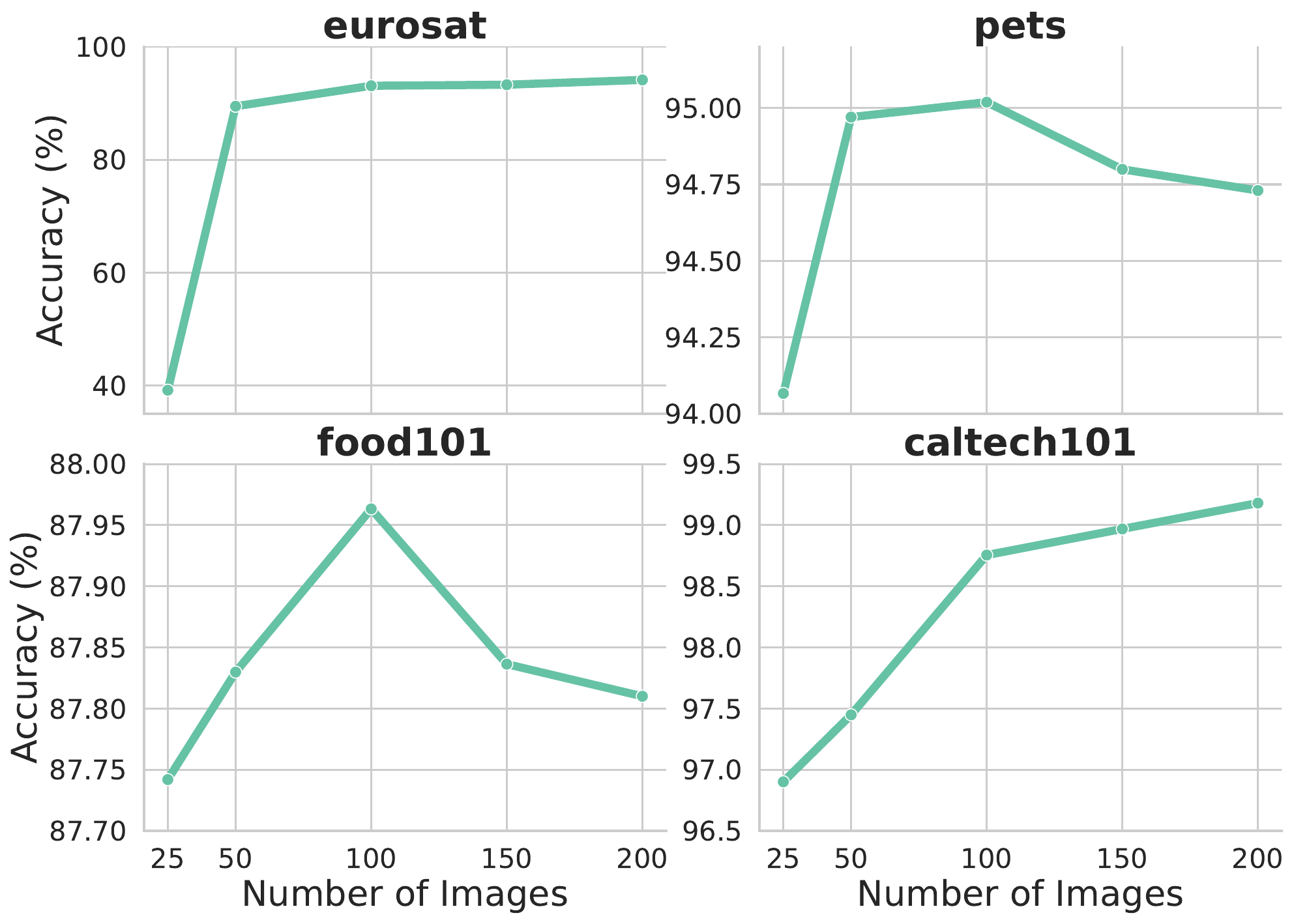}
    \caption{
      Varying the number of generated images per class.
    }
    \label{fig:acc_vs_n_img}
  \end{minipage}
  \hfill
  \begin{minipage}[t]{0.48\linewidth}
    \centering
    \includegraphics[width=\linewidth]{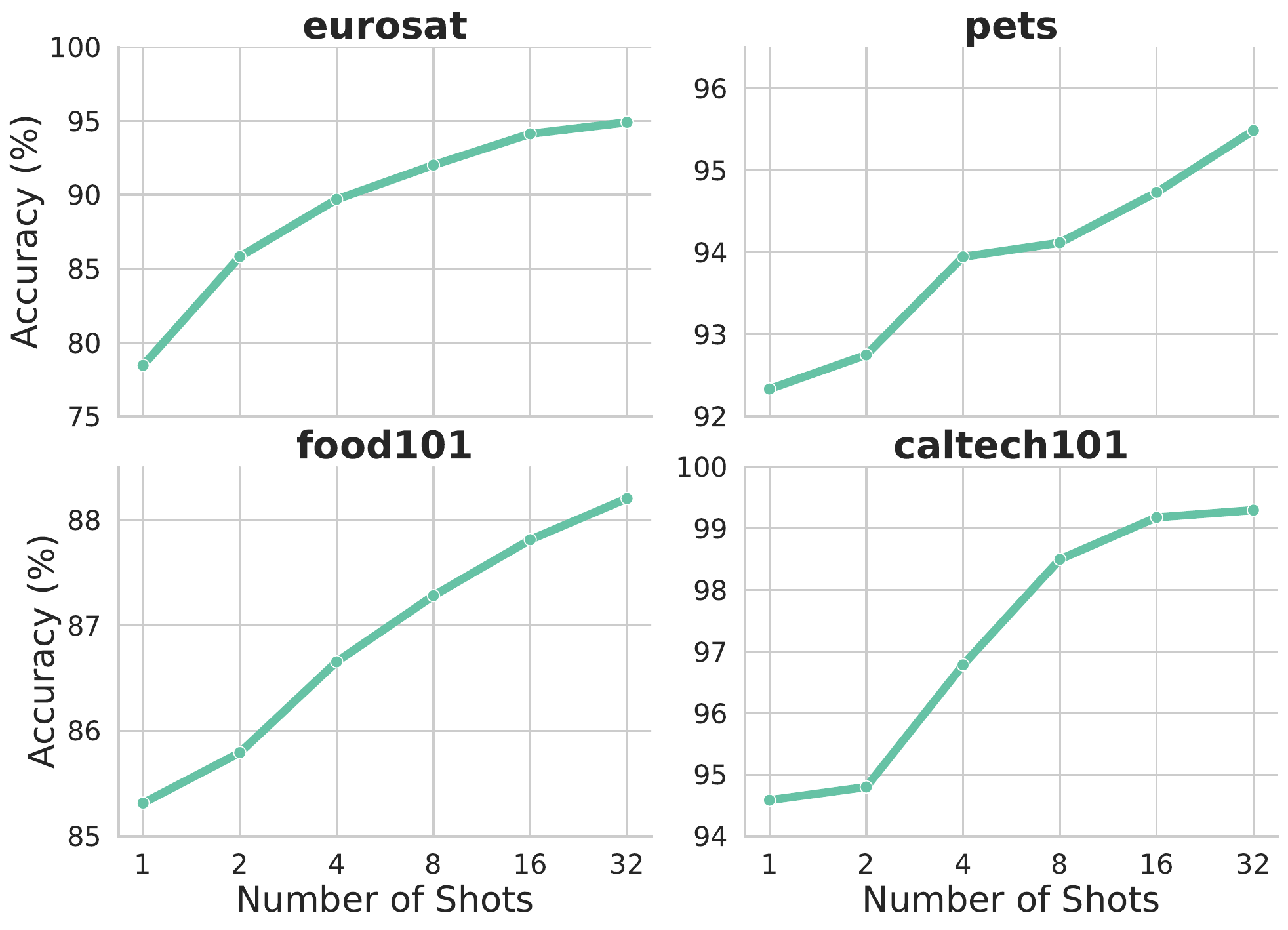}
    \caption{
      Varying the number of real few-shot images per class. 
    }
    \label{fig:acc_vs_n_shots}
  \end{minipage}

\end{figure*}

\paragraph{Ablations of the number of generated and the few-shot real images.}
We study how the number of generated and few-shot real images affects performance (Figs. \ref{fig:acc_vs_n_img} and \ref{fig:acc_vs_n_shots}). Accuracy generally improves with more generated data, with 200 images per class providing a good balance of performance and computational cost.
For Food101, the accuracy drop beyond 100 images is likely due to its high intra-class variability, where extra samples deviate from the real test distribution, leading the classifier to overfit synthetic-specific cues. %
We then test the impact of the number of real images used for generation. While even one shot yields acceptable accuracy, performance improves with more examples, and 16 shots offer a strong trade-off between data efficiency and generalization.

\section{Conclusion}
We proposed DIPSY, a novel training-free approach to few-shot image classification through dual image prompting with positive-negative guidance. By extending classifier-free guidance for independent control of image conditioning and implementing a class similarity-based sampling strategy, our method achieves competitive performance across ten benchmark datasets without requiring model fine-tuning, captioning, or filtering as modern methods do. 
Future work includes extending DIPSY to video and 3D domains, and applying it to more complex tasks, such as few-shot object detection and segmentation.

\section*{Acknowledgements}
This project was partially supported by a Hi!Paris collaborative project, the Agence de l'Innovation de Défense – AID - via Centre Interdisciplinaire d’Etudes pour la Défense et la Sécurité – CIEDS - (project 2024 - FakeDetect) and was granted access to the HPC resources of IDRIS under the allocation 2024-AD011015893 made by GENCI. The authors would like to thank Xi Wang for valuable discussions, Guillaume Astruc for proofreading and the anonymous reviewers for their insightful comments and suggestions.

\bibliography{shortstrings, egbib}

\clearpage
\appendix
\section*{Supplementary Material}
This appendix provides supplementary materials for our paper. We include: (1) a detailed pseudocode for our proposed \mname{} synthetic image generation process and (2) additional qualitative results demonstrating the visual fidelity of generated images across various datasets and classes. These materials aim to offer greater insight into our methodology.

\section{Algorithm Details}
\label{sec:algorithm}

The core of our proposed \mname{} method involves a pipeline for generating synthetic images that are both class-representative and discriminative. This process leverages dual IP-Adapter guidance, conditioning on a positive image prompt from the target class and a negative image prompt from a similar class, alongside a text prompt. Alg.~\ref{alg:dipsy_generation_simple} provides a step-by-step breakdown of this generation process.

\begin{algorithm}
\caption{DIPSY: Core Synthetic Image Generation}
\label{alg:dipsy_generation_simple}
\begingroup
\ttfamily
\footnotesize
\begin{algorithmic}[1]
\Procedure{DIPSYGeneration}{RealFewShotDataset, $n_{synth}$, $w_{\text{text}}, w_{\text{im+}}, w_{\text{im-}}$}
    \State \textbf{Input:}
    \State RealFewShotDataset: Small set of labeled real images for each class.
    \State $n_{synth}$: Number of synthetic images to create per class.
    \State $w_{\text{text}}$: Guidance scale for the text prompt.
    \State $w_{\text{im+}}$: Guidance scale for the positive image prompt.
    \State $w_{\text{im-}}$: Guidance scale for the negative image prompt. \medskip
    \State \textbf{Output:} SyntheticDataset \medskip
    
    \State SyntheticDataset $\gets \emptyset$
    
    \ForAll{class $C_{+}$ in RealFewShotDataset}
        \For{$i = 1$ \textbf{to} $n_{synth}$}
            \State \textit{// Step 1: Select Guiding Images from Few-Shot Examples}
            \State $I_{\text{+}} \gets$ Randomly pick one real image from $C_{+}$.
            \State $C_{-} \gets$ Probabilistically sample a class from AllClasses (excluding $C_{+}$)
            \Statex \hspace{7em} where classes more similar to $C_{+}$ are more likely to be chosen,
            \Statex \hspace{7em} based on CLIP similarity of few-shot images
            \State $I_{\text{-}} \gets$ Randomly pick one real image from $C_{-}$. \medskip

            \State \textit{// Step 2: Prepare Prompts and Augment Images}
            \State TextPrompt $\gets$ "A photo of a " + name of $C_{+}$ \textit{(dataset-specific formulation)}.
            \State $I_{\text{+}} \gets$ Augment($I_{\text{+}}$) \Comment{e.g., random crop, rotation}
            \State $I_{\text{-}} \gets$ Augment($I_{\text{-}}$) \medskip

            \State \textit{// Step 3: Generate Image using Dual IP-Adapter Guidance}
            \State NewImage $\gets$ DiffusionModel.Generate(
            \Statex \hspace{5em} text\_condition $\gets$ TextPrompt,
            \Statex \hspace{5em} positive\_image\_condition $\gets I_{\text{+}}$,
            \Statex \hspace{5em} negative\_image\_condition $\gets I_{\text{-}}$,
            \Statex \hspace{5em} text\_guidance\_scale $\gets w_{\text{text}}$,
            \Statex \hspace{5em} positive\_image\_guidance\_scale $\gets w_{\text{im+}}$,
            \Statex \hspace{5em} negative\_image\_guidance\_scale $\gets w_{\text{im-}}$
            \Statex \hspace{4.5em})
            
            \State \medskip Add (NewImage, $C_{+}$) to SyntheticDataset.
        \EndFor
    \EndFor
    \State \Return SyntheticDataset
\EndProcedure

\end{algorithmic}
\endgroup
\end{algorithm}

\section{Additional Qualitative Results}
\label{sec:qualitative}

To further illustrate the visual fidelity and class-specificity of images generated by DIPSY, this section presents additional qualitative examples across diverse fine-grained categories, such as Pets and Flowers. As can be observed in the comparisons (Figs.~\ref{fig:grouped_by_method_pets} and \ref{fig:grouped_by_method_flower}), the synthetic images demonstrate a high degree of realism, closely resembling genuine photographs from the respective datasets. DIPSY effectively captures the defining visual characteristics and intricate details specific to each target class. For instance, pet breeds exhibit their distinct features and fur patterns, and flower varieties display appropriate petal structures and natural coloration. Moreover, the generated images generally exhibit an overall aesthetic distribution consistent with the original, real-world examples for each class. These examples show DIPSY's capability to synthesize visually plausible and class-consistent data, suitable for augmenting training sets in few-shot learning scenarios.

\begin{table*}[h]
\begin{minipage}[t]{\textwidth}

  \centering
  \setlength{\tabcolsep}{2pt}
  \resizebox{\linewidth}{!}{%
  \begin{tabular}{cc}
    \begin{tabular}{c}
      \textbf{Real Images} \\[1mm]
      \textit{English Cocker Spaniel} \\
      \includegraphics[valign=m,width=0.22\linewidth]{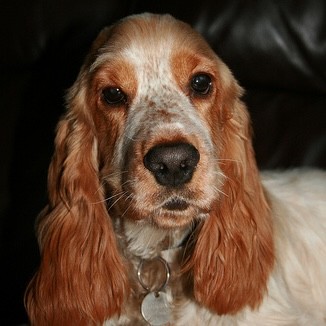} 
      \includegraphics[valign=m,width=0.22\linewidth]{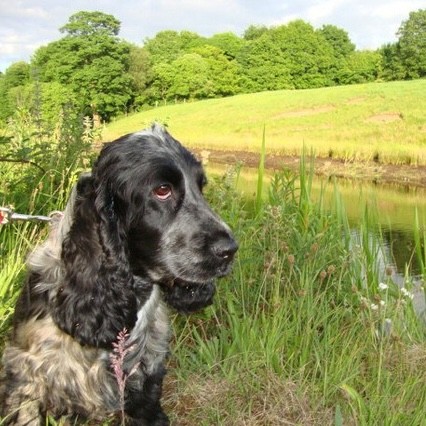} 
      \includegraphics[valign=m,width=0.22\linewidth]{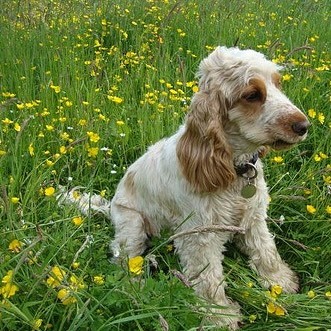} 
      \includegraphics[valign=m,width=0.22\linewidth]{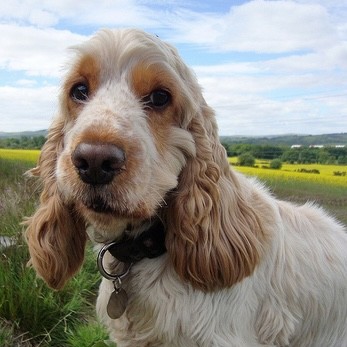} \\
      \textit{Persian} \\
      \includegraphics[valign=m,width=0.22\linewidth]{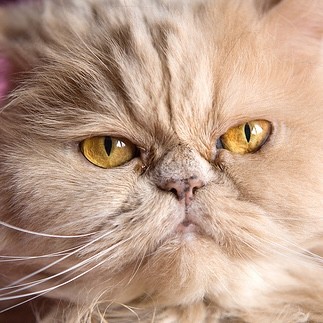} 
      \includegraphics[valign=m,width=0.22\linewidth]{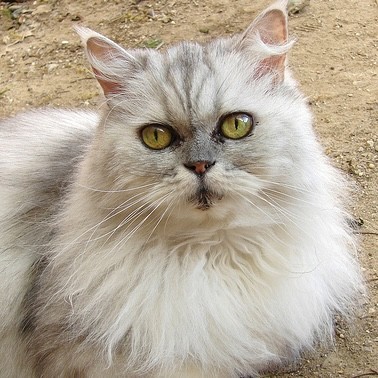} 
      \includegraphics[valign=m,width=0.22\linewidth]{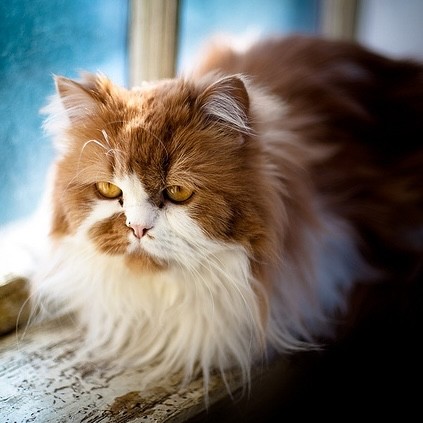} 
      \includegraphics[valign=m,width=0.22\linewidth]{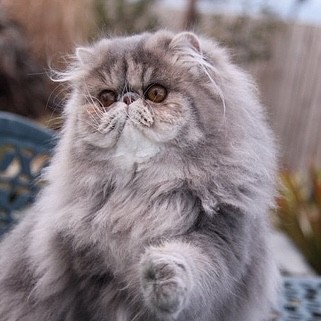} \\
      \textit{Beagle} \\
      \includegraphics[valign=m,width=0.22\linewidth]{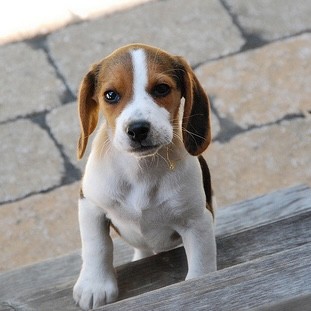} 
      \includegraphics[valign=m,width=0.22\linewidth]{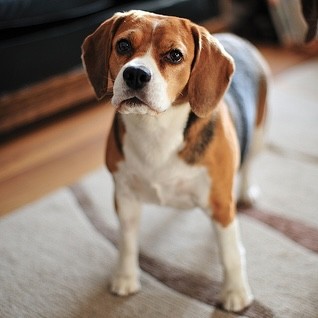} 
      \includegraphics[valign=m,width=0.22\linewidth]{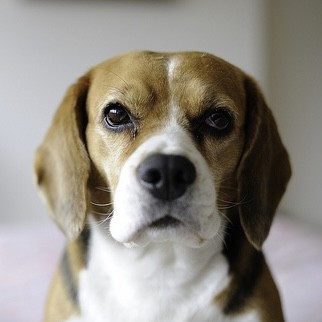} 
      \includegraphics[valign=m,width=0.22\linewidth]{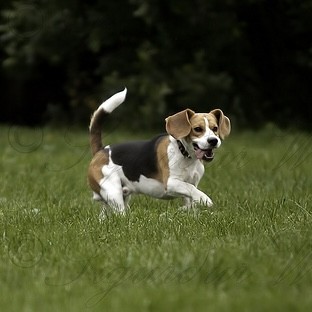} \\
    \end{tabular}
    &
    \begin{tabular}{c}
      \textbf{DIPSY (ours)} \\[1mm]
      \textit{English Cocker Spaniel} \\
      \includegraphics[valign=m,width=0.22\linewidth]{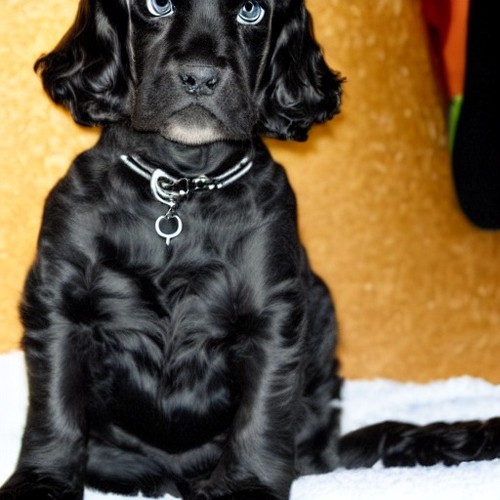} 
      \includegraphics[valign=m,width=0.22\linewidth]{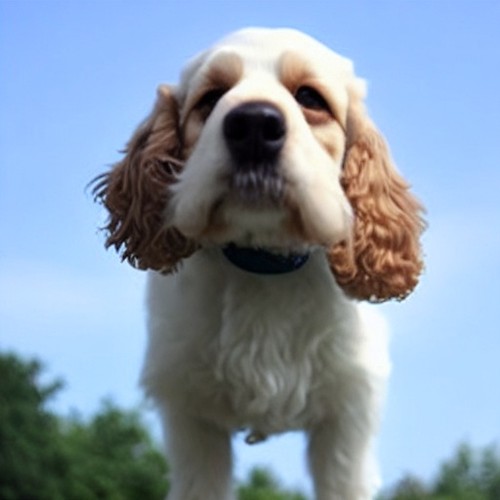} 
      \includegraphics[valign=m,width=0.22\linewidth]{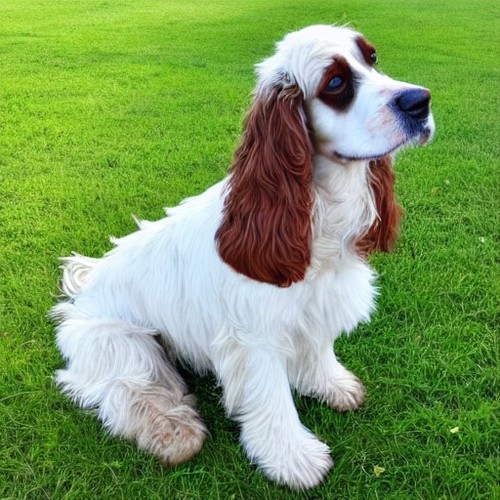} 
      \includegraphics[valign=m,width=0.22\linewidth]{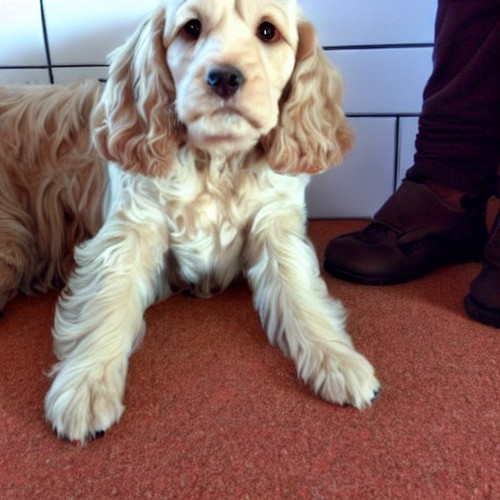} \\
      \textit{Persian} \\
      \includegraphics[valign=m,width=0.22\linewidth]{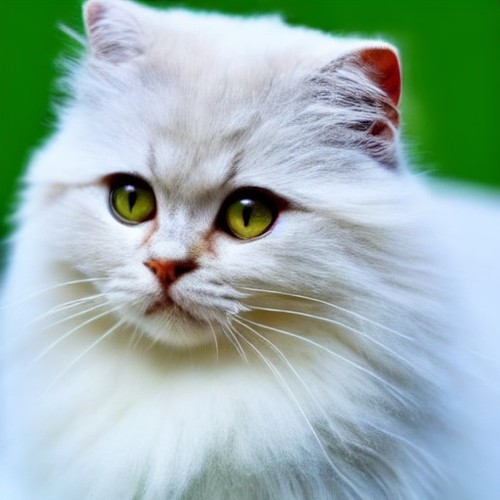} 
      \includegraphics[valign=m,width=0.22\linewidth]{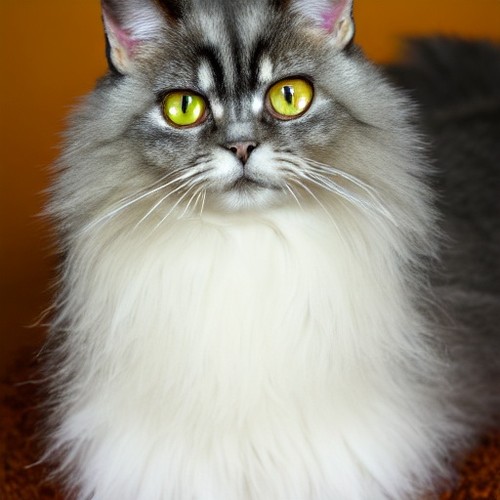} 
      \includegraphics[valign=m,width=0.22\linewidth]{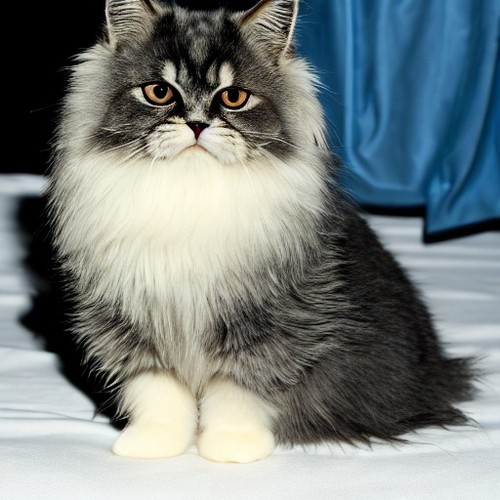} 
      \includegraphics[valign=m,width=0.22\linewidth]{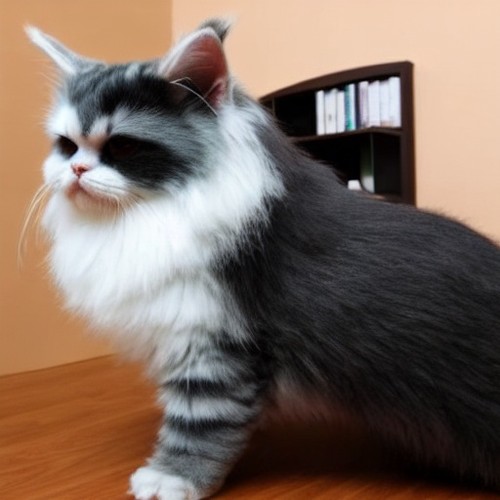} \\
      \textit{Beagle} \\
      \includegraphics[valign=m,width=0.22\linewidth]{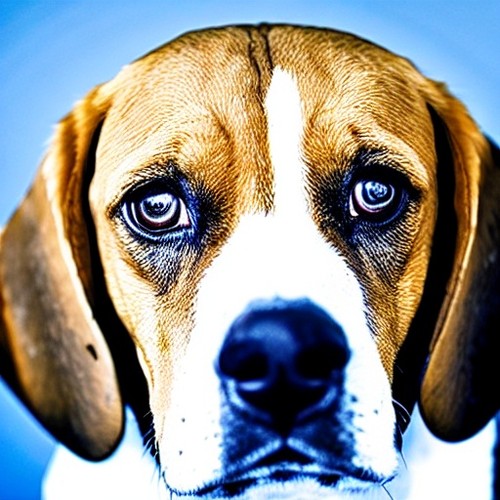} 
      \includegraphics[valign=m,width=0.22\linewidth]{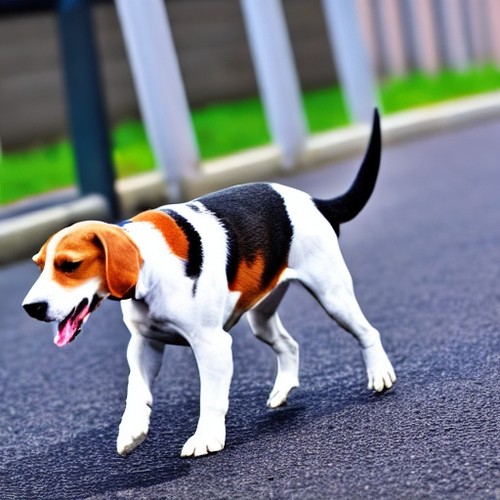} 
      \includegraphics[valign=m,width=0.22\linewidth]{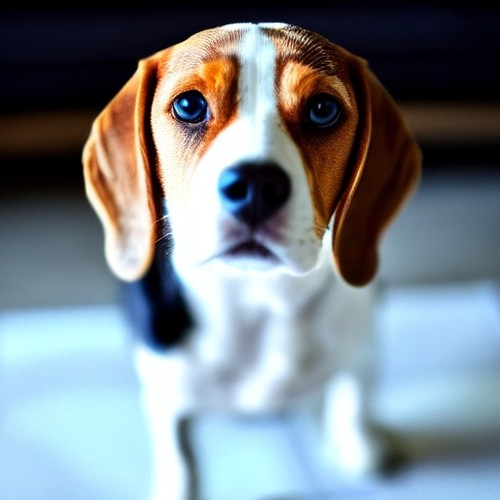} 
      \includegraphics[valign=m,width=0.22\linewidth]{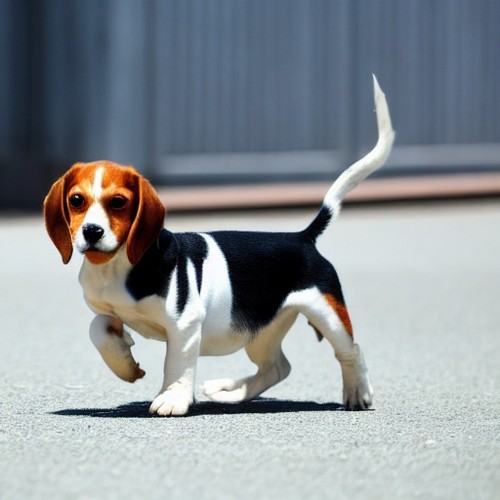} \\
    \end{tabular}
    \\
  \end{tabular}}
  \vspace{5mm}
  \captionof{figure}{
    Qualitative comparison of real and DIPSY generated synthetic images from the classes English Cocker Spaniel, Persian and Beagle from the Oxford-IIIT Pet dataset. \mname{} generates realistic, class-specific images that capture breed characteristics like coat patterns and facial structures.
  }
  \label{fig:grouped_by_method_pets}
\end{minipage}
\begin{minipage}[t]{\textwidth}
  \centering
  \setlength{\tabcolsep}{2pt}
  \resizebox{\linewidth}{!}{%
  \begin{tabular}{cc}
    \begin{tabular}{c}
      \textbf{Real Images} \\[1mm]
      \textit{Siam Tulip} \\
      \includegraphics[valign=m,width=0.22\linewidth]{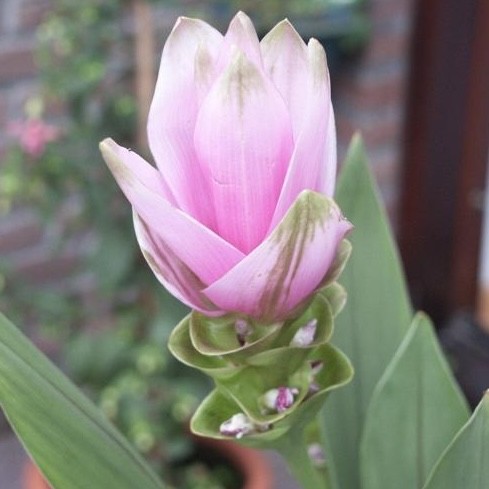} 
      \includegraphics[valign=m,width=0.22\linewidth]{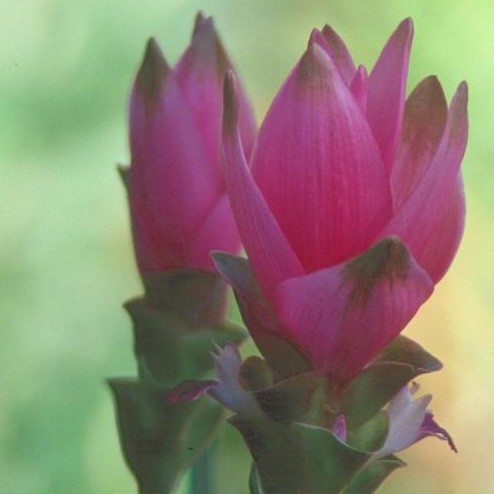} 
      \includegraphics[valign=m,width=0.22\linewidth]{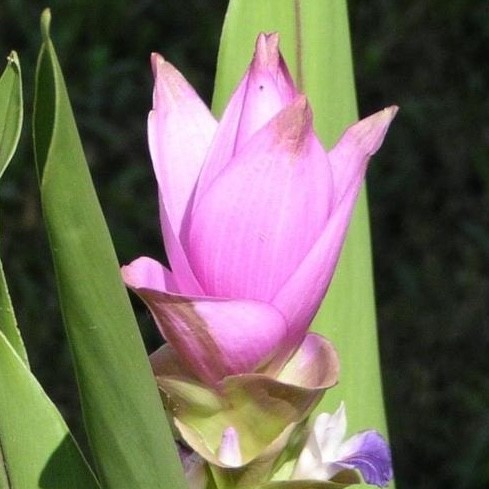} 
      \includegraphics[valign=m,width=0.22\linewidth]{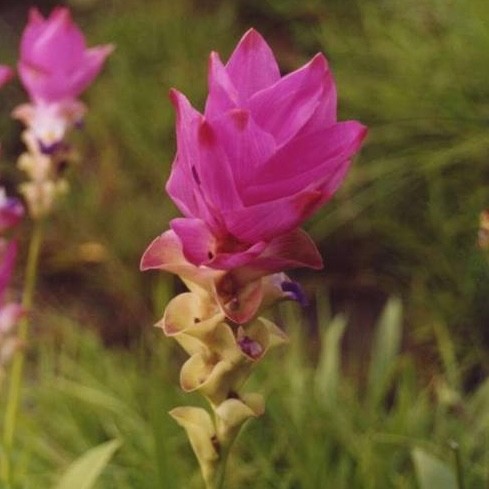} \\
      \textit{Rose} \\
      \includegraphics[valign=m,width=0.22\linewidth]{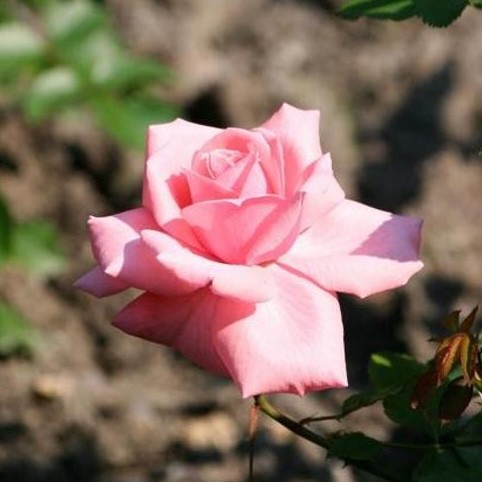} 
      \includegraphics[valign=m,width=0.22\linewidth]{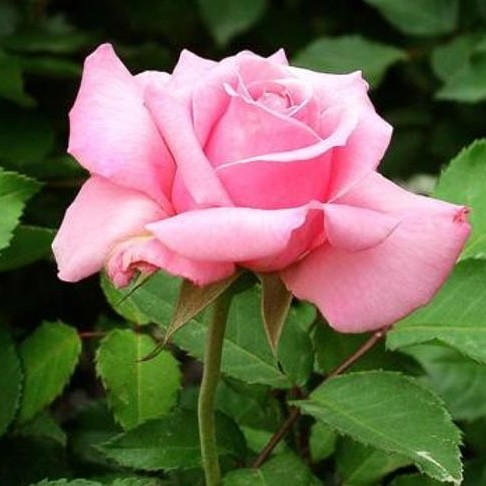} 
      \includegraphics[valign=m,width=0.22\linewidth]{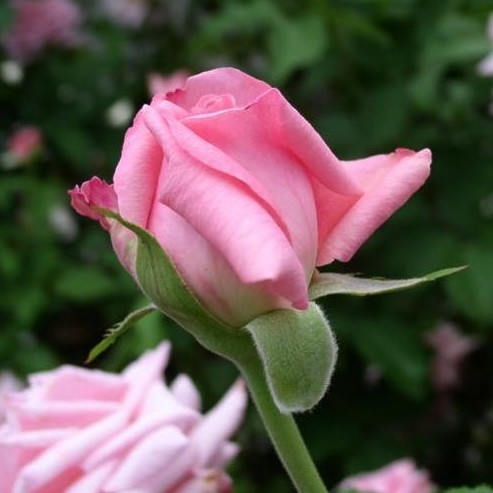} 
      \includegraphics[valign=m,width=0.22\linewidth]{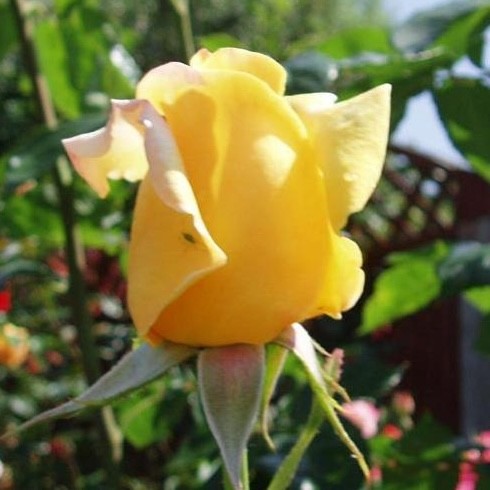} \\
      \textit{Yellow Iris} \\
      \includegraphics[valign=m,width=0.22\linewidth]{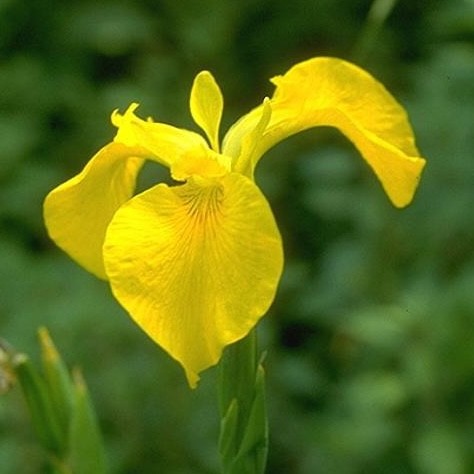} 
      \includegraphics[valign=m,width=0.22\linewidth]{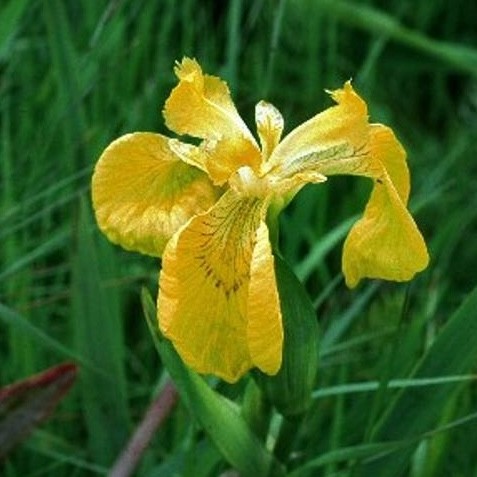} 
      \includegraphics[valign=m,width=0.22\linewidth]{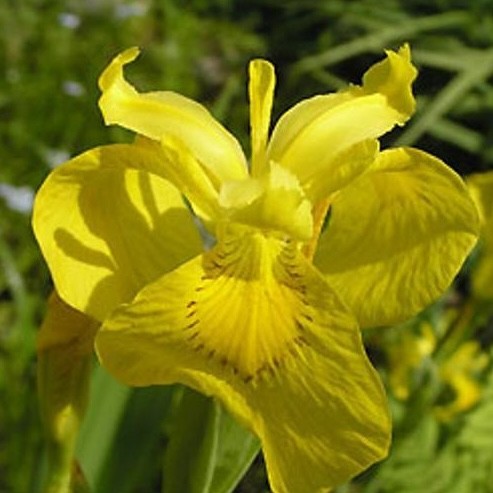} 
      \includegraphics[valign=m,width=0.22\linewidth]{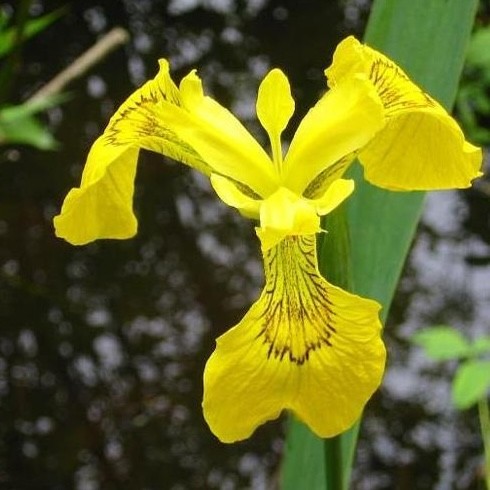} \\
    \end{tabular}
    &
    \begin{tabular}{c}
      \textbf{DIPSY (ours)} \\[1mm]
      \textit{Siam Tulip} \\
      \includegraphics[valign=m,width=0.22\linewidth]{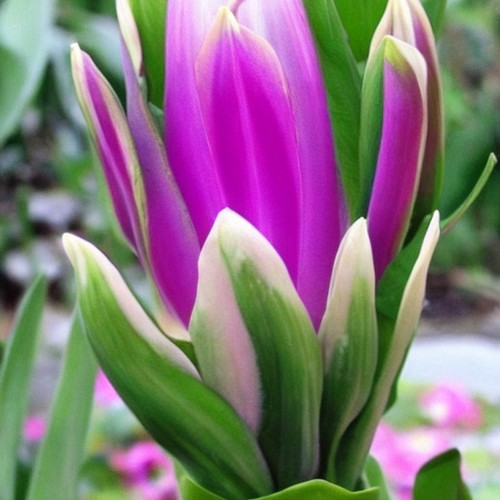} 
      \includegraphics[valign=m,width=0.22\linewidth]{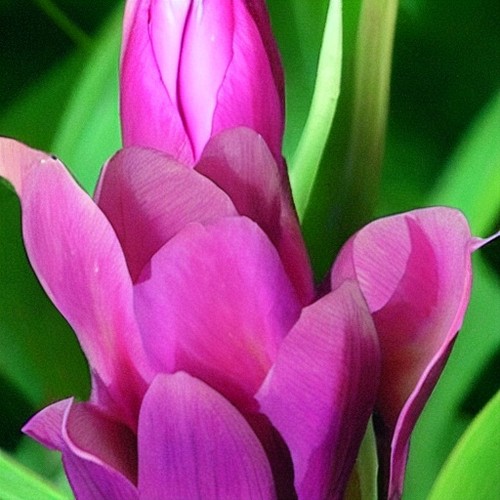} 
      \includegraphics[valign=m,width=0.22\linewidth]{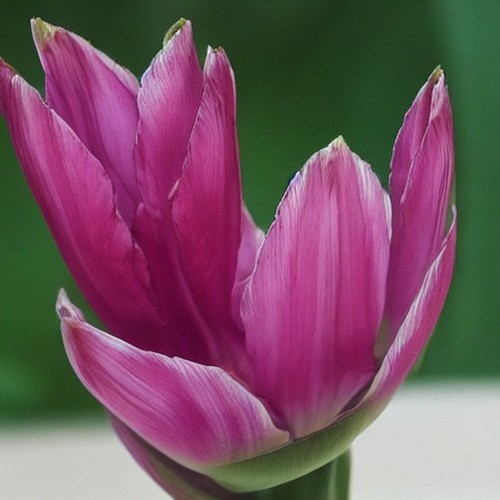} 
      \includegraphics[valign=m,width=0.22\linewidth]{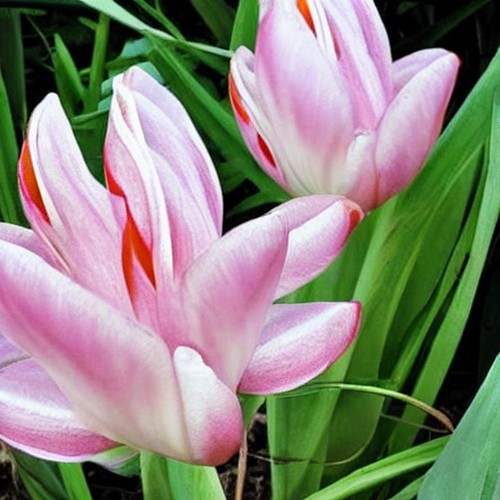} \\
      \textit{Rose} \\
      \includegraphics[valign=m,width=0.22\linewidth]{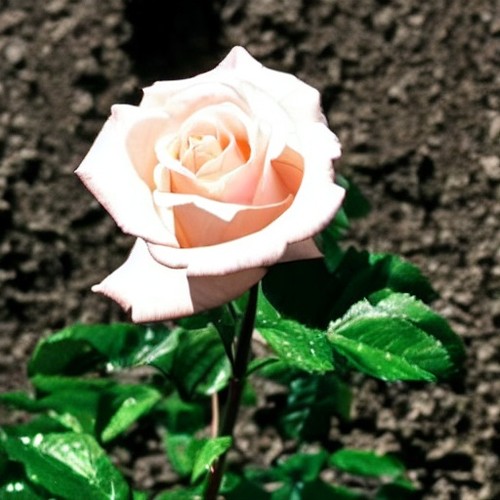} 
      \includegraphics[valign=m,width=0.22\linewidth]{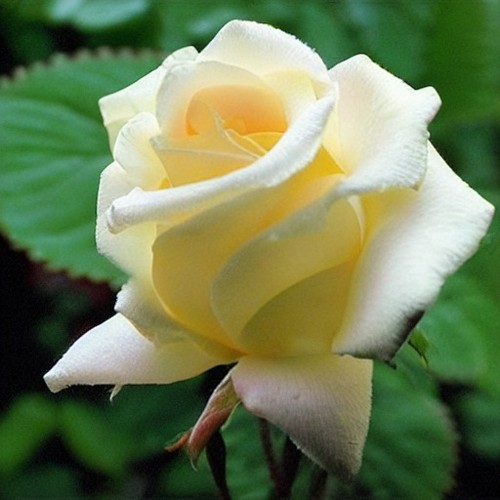} 
      \includegraphics[valign=m,width=0.22\linewidth]{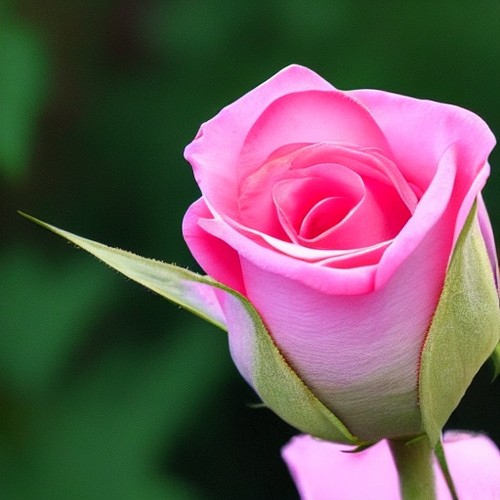} 
      \includegraphics[valign=m,width=0.22\linewidth]{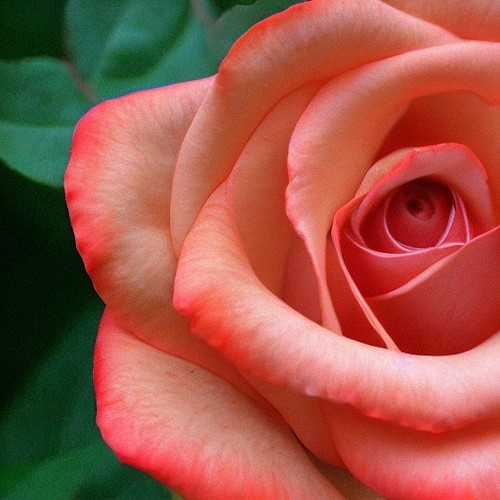} \\
      \textit{Yellow Iris} \\
      \includegraphics[valign=m,width=0.22\linewidth]{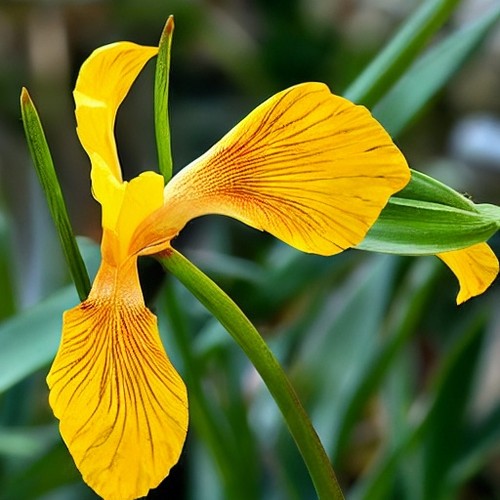} 
      \includegraphics[valign=m,width=0.22\linewidth]{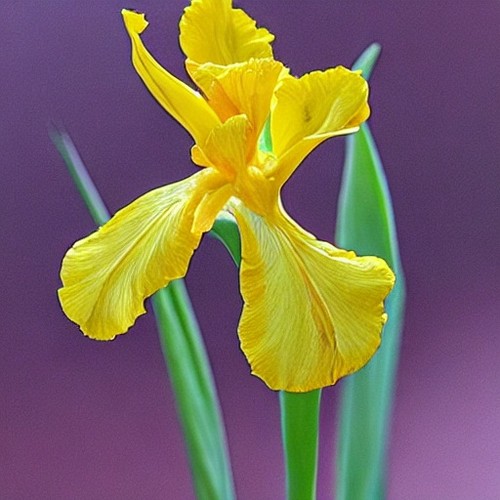} 
      \includegraphics[valign=m,width=0.22\linewidth]{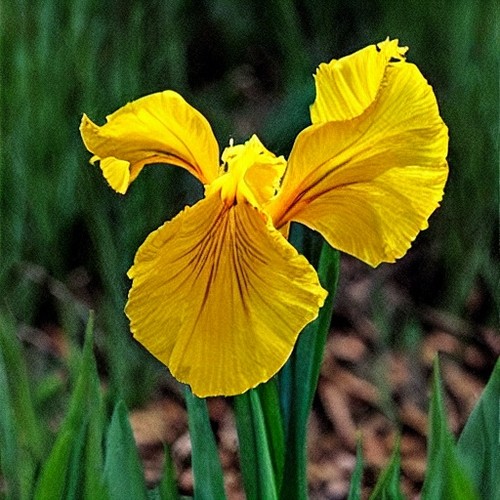} 
      \includegraphics[valign=m,width=0.22\linewidth]{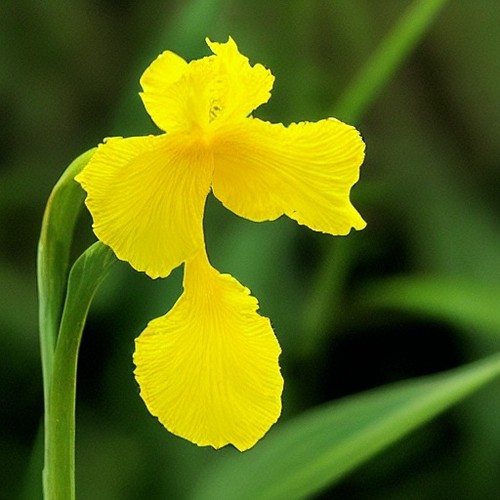} \\
    \end{tabular}
    \\
  \end{tabular}}
  \vspace{5mm}
  \captionof{figure}{
    Qualitative comparison of real and DIPSY generated synthetic images from the classes Siam Tulip, Rose and Yellow Iris from the Oxford Flowers 102 dataset. The \mname{} generated images accurately represent the unique features, colors, and forms of each flower class, closely resembling the real images.
  }
  \label{fig:grouped_by_method_flower}
\end{minipage}
\end{table*}

\end{document}